\pdfoutput=1
\documentclass[10pt,twocolumn,letterpaper]{article}

\usepackage[pagenumbers]{cvpr} 

\usepackage{graphicx}
\usepackage{amsmath}
\usepackage{amssymb}
\usepackage{booktabs}
\usepackage{lipsum}
\usepackage{booktabs}
\usepackage{bbding}
\usepackage{url}
\usepackage{multirow}
\usepackage[accsupp]{axessibility}  

\usepackage[pagebackref,breaklinks,colorlinks]{hyperref}

\usepackage{color}
\usepackage{enumitem}
\definecolor{mypink3}{cmyk}{0, 0.7808, 0.4429, 0.1412}
\definecolor{myblue2}{cmyk}{0, 0., 0.4412, 0.4808}

\definecolor{myblue}{cmyk}{0, 0.7808, 0., 0.1412}

\usepackage[capitalize]{cleveref}
\crefname{section}{Sec.}{Secs.}
\Crefname{section}{Section}{Sections}
\Crefname{table}{Table}{Tables}
\crefname{table}{Tab.}{Tabs.}

\def\cvprPaperID{623} 
\def\confName{CVPR}
\def\confYear{2022}

\begin{document}

\title{Stratified Transformer for 3D Point Cloud Segmentation}


\author{Xin Lai$^{1}$\thanks{Equal Contribution}\hspace{1.0cm}Jianhui Liu$^{2,3*}$\hspace{1.0cm}Li Jiang$^{4}$\hspace{1.0cm}Liwei Wang$^{1}$\hspace{1.0cm}Hengshuang Zhao$^{2,5}$\\Shu Liu$^{3}$\hspace{1.0cm}Xiaojuan Qi$^{2}$\thanks{Corresponding Author}\hspace{1.0cm}Jiaya Jia$^{1,3}$\\
$^{1}$CUHK~~~
$^{2}$HKU~~~
$^{3}$SmartMore~~~
$^{4}$MPI Informatics~~~
$^{5}$MIT\\
}
\maketitle

\begin{abstract}
3D point cloud segmentation has made tremendous progress in recent years. Most current methods focus on aggregating local features, but fail to directly model long-range dependencies. In this paper, we propose \textbf{Stratified Transformer} that is able to capture long-range contexts and demonstrates strong generalization ability and high performance. Specifically, we first put forward a novel key sampling strategy. For each query point, we sample nearby points densely and distant points sparsely as its keys in a stratified way, which enables the model to enlarge the effective receptive field and enjoy long-range contexts at a low computational cost. Also, to combat the challenges posed by irregular point arrangements, we propose first-layer point embedding to aggregate local information, which facilitates convergence and boosts performance. Besides, we adopt contextual relative position encoding to adaptively capture position information. Finally, a memory-efficient implementation is introduced to overcome the issue of varying point numbers in each window. Extensive experiments demonstrate the effectiveness and superiority of our method on S3DIS, ScanNetv2 and ShapeNetPart datasets. Code is available at \url{https://github.com/dvlab-research/Stratified-Transformer}.
\end{abstract}

\section{Introduction}
\label{sec:intro}

Nowadays 3D point clouds can be conveniently collected. They have demonstrated great potential in various applications, such as autonomous driving, robotics and augmented reality. Unlike regular pixels in 2D images, 3D points are arranged irregularly
, hampering direct adoption of well-studied 2D networks to process 3D data. Therefore, it is imperative to explore advanced methods that are tailored for 3D point cloud data.

Abundant methods~\cite{qi2017pointnet, qi2017pointnet++, wu2019pointconv, thomas2019kpconv, zhao2020point, zhao2019pointweb, 3DSemanticSegmentationWithSubmanifoldSparseConvNet, SubmanifoldSparseConvNet, choy20194d} have explored 3D point cloud segmentation and obtained decent performance. Most of them focus on aggregating local features, but fail to explicitly model long-range dependencies, which has been demonstrated to be crucial in capturing contexts from a long distance~\cite{wang2018non}. Along another line of research, Transformer~\cite{vaswani2017attention} can naturally harvest long-range information via the self-attention mechanism. However, only limited attempts~\cite{zhao2020point, mao2021voxel} have been made to apply Transformer to 3D point clouds. Point Transformer~\cite{zhao2020point} proposes ``vector self-attention" and ``subtraction relation" to aggregate local features, but it is still difficult to directly capture long-range contexts. 
Voxel Transformer~\cite{mao2021voxel} is tailored for object detection and performs self-attention over the voxels, but it loses accurate position due to voxelization. 

Differently, we develop an efficient segmentation network to capture long-range contexts using the standard multi-head self-attention~\cite{vaswani2017attention}, while keeping position information intact.
To this end, we propose a simple and powerful framework, namely, \textit{Stratified Transformer}.


\begin{figure}
\begin{center}

	\centering
    \begin{minipage}  {0.32\linewidth}
        \centering
        \includegraphics [width=1\linewidth,height=0.6\linewidth]
        {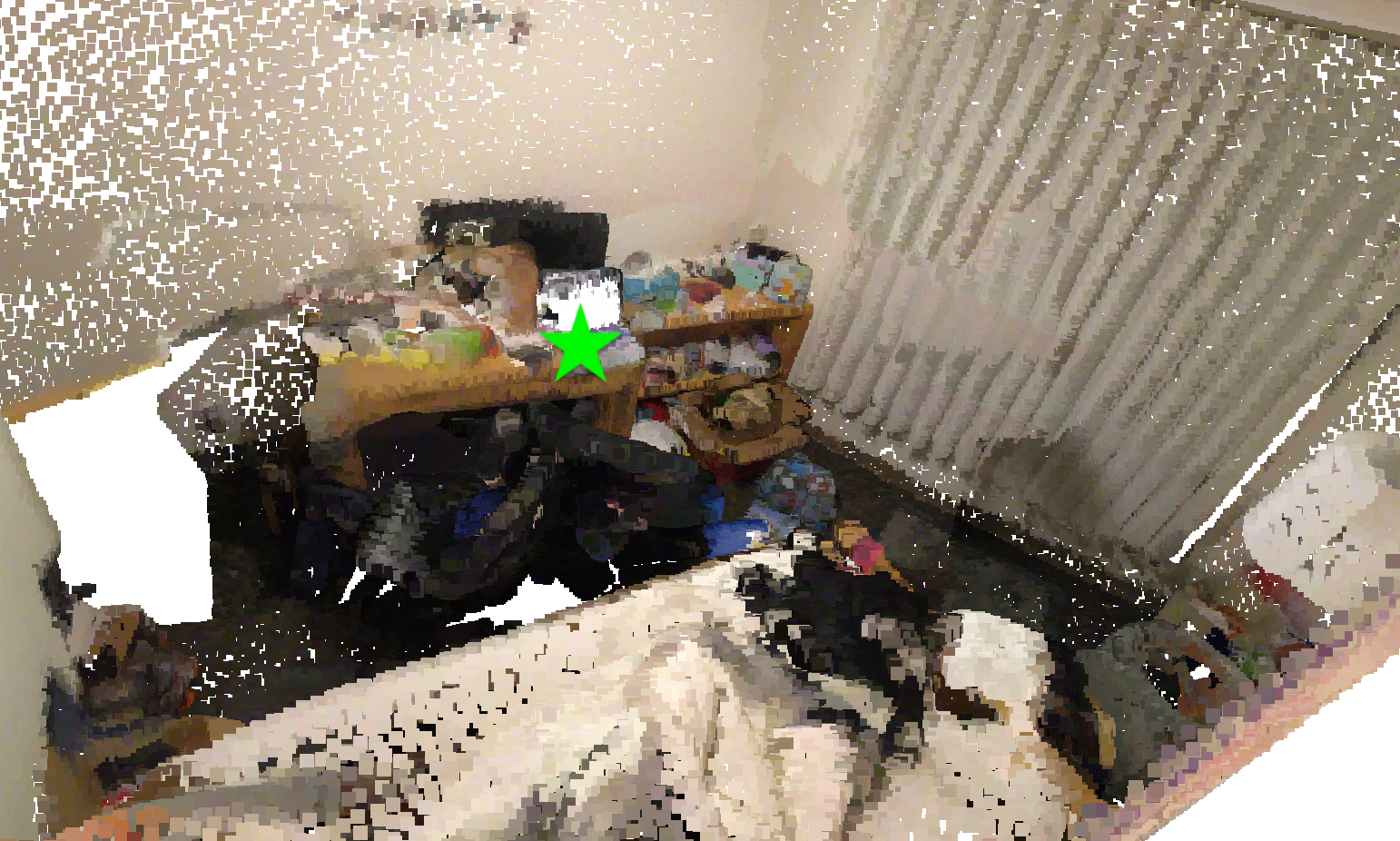}
    \end{minipage}      
    \begin{minipage}  {0.32\linewidth}
        \centering
        \includegraphics [width=1\linewidth,height=0.6\linewidth]
        {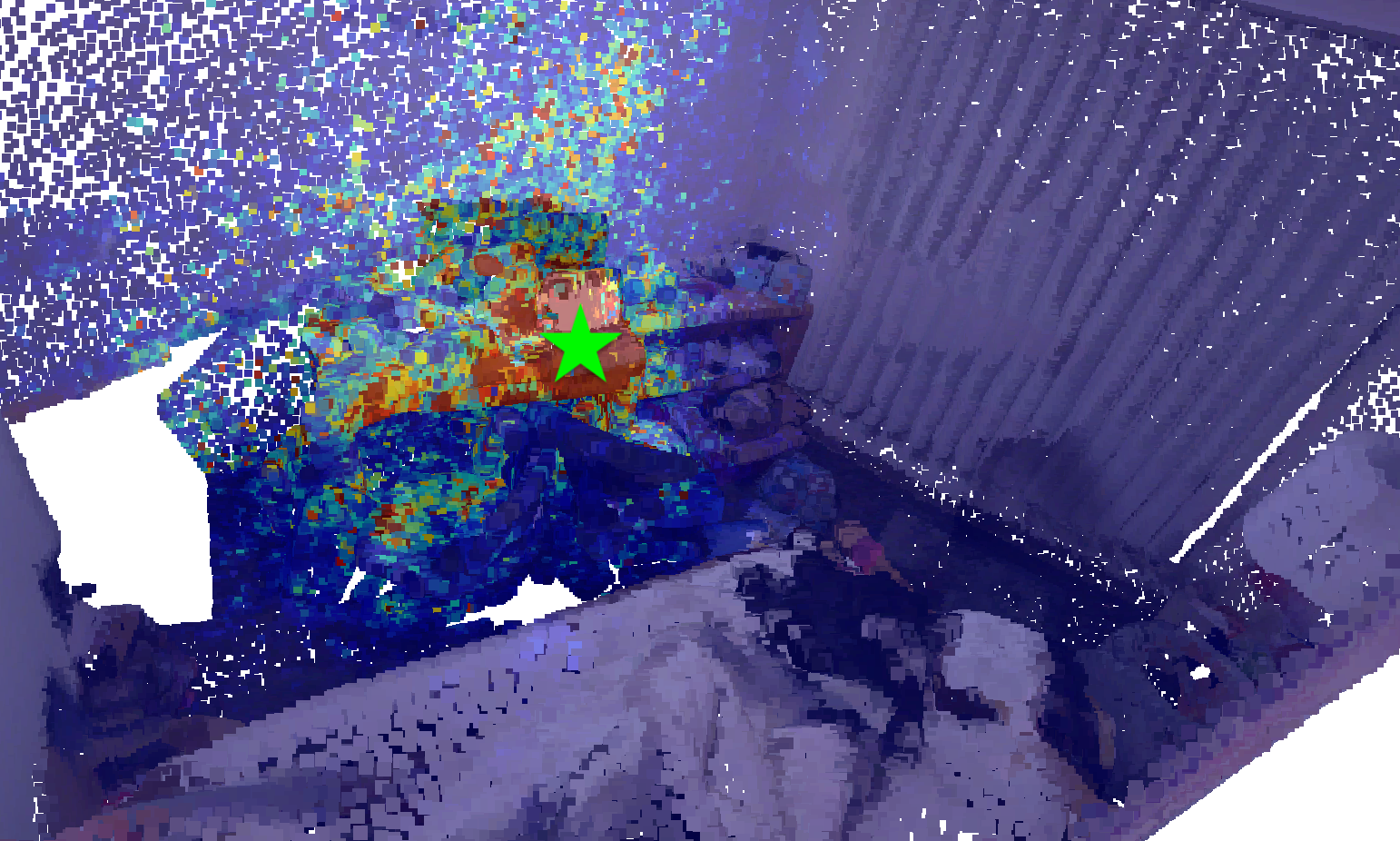}
    \end{minipage}      
     \begin{minipage}  {0.32\linewidth}
        \centering
        \includegraphics [width=1\linewidth,height=0.6\linewidth]
        {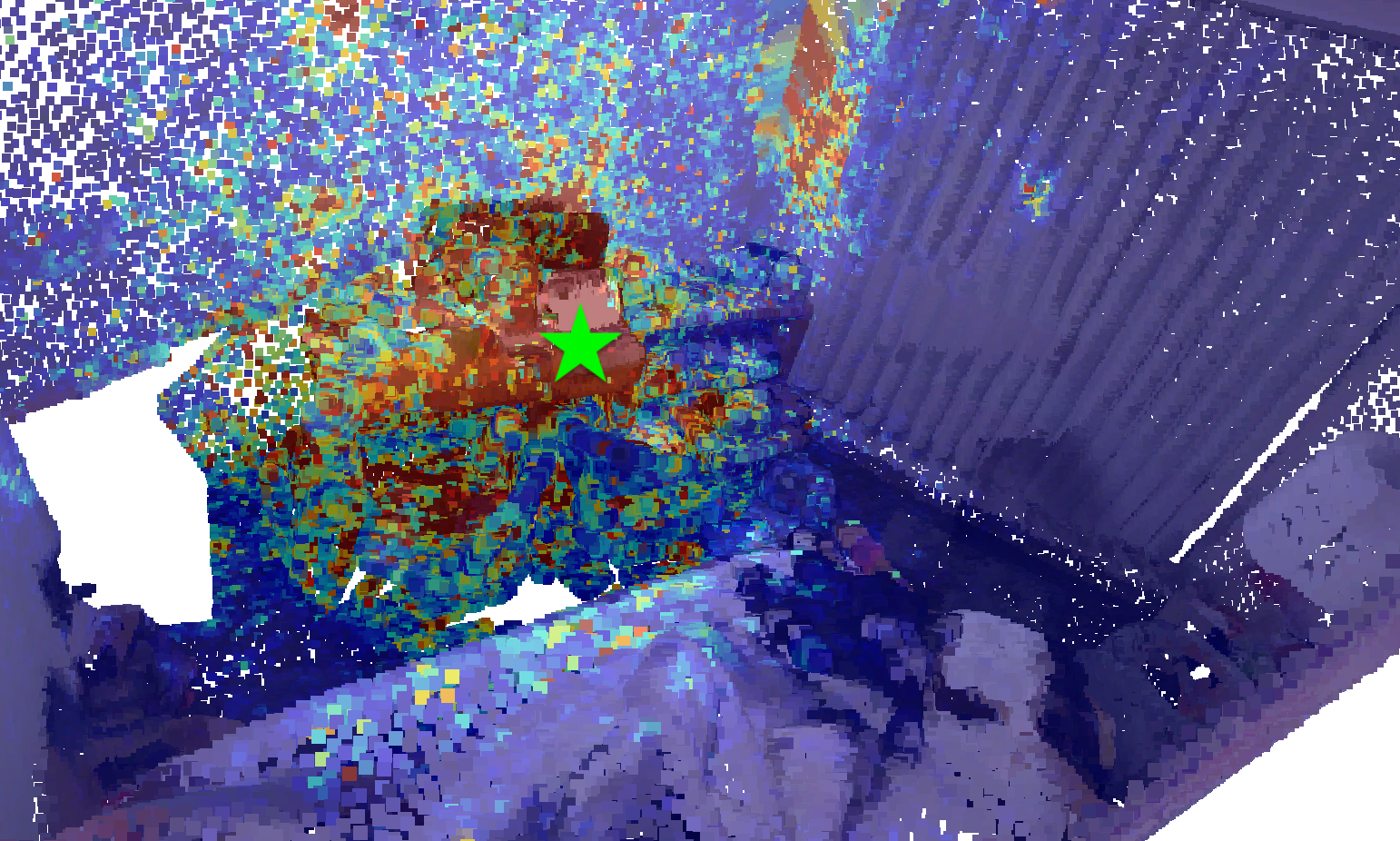}
    \end{minipage} 
	 
	 
    \begin{minipage}  {0.32\linewidth}
        \centering
        \includegraphics [width=1\linewidth,height=0.6\linewidth]
        {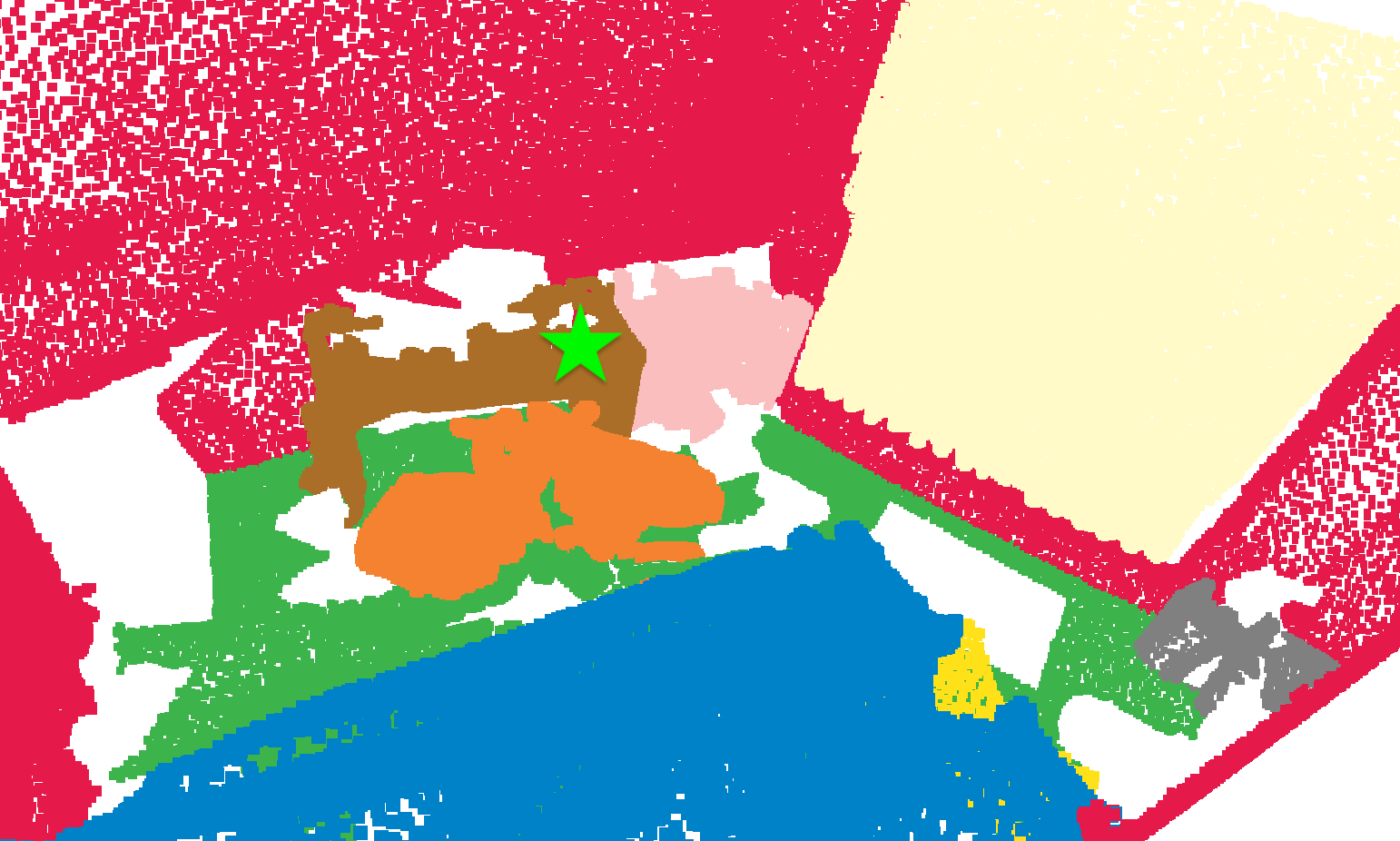}\\\footnotesize Input / Ground Truth
    \end{minipage}      
    \begin{minipage}  {0.32\linewidth}
        \centering
        \includegraphics [width=1\linewidth,height=0.6\linewidth]
        {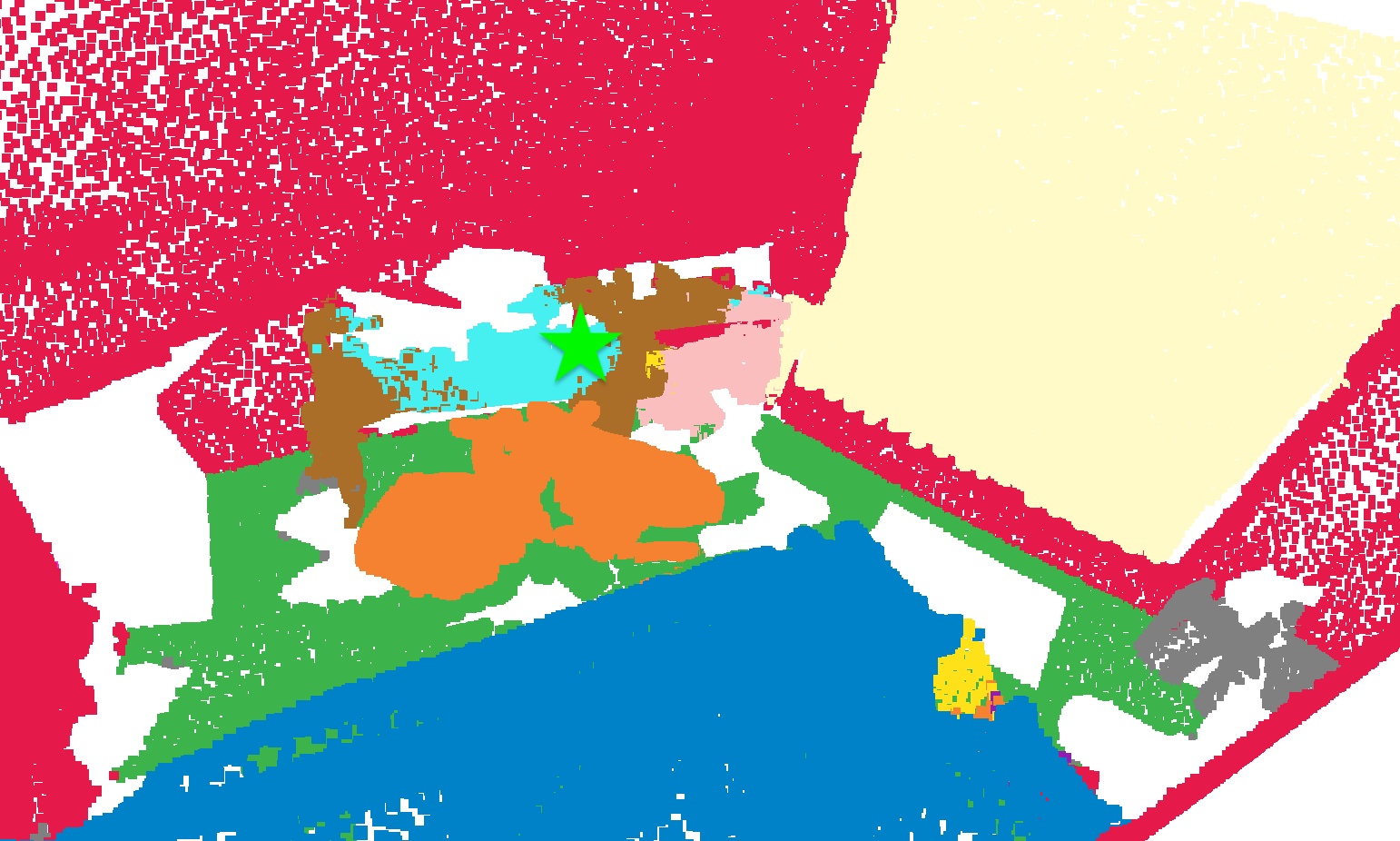}\\\footnotesize w/o stratified
    \end{minipage}      
     \begin{minipage}  {0.32\linewidth}
        \centering
        \includegraphics [width=1\linewidth,height=0.6\linewidth]
        {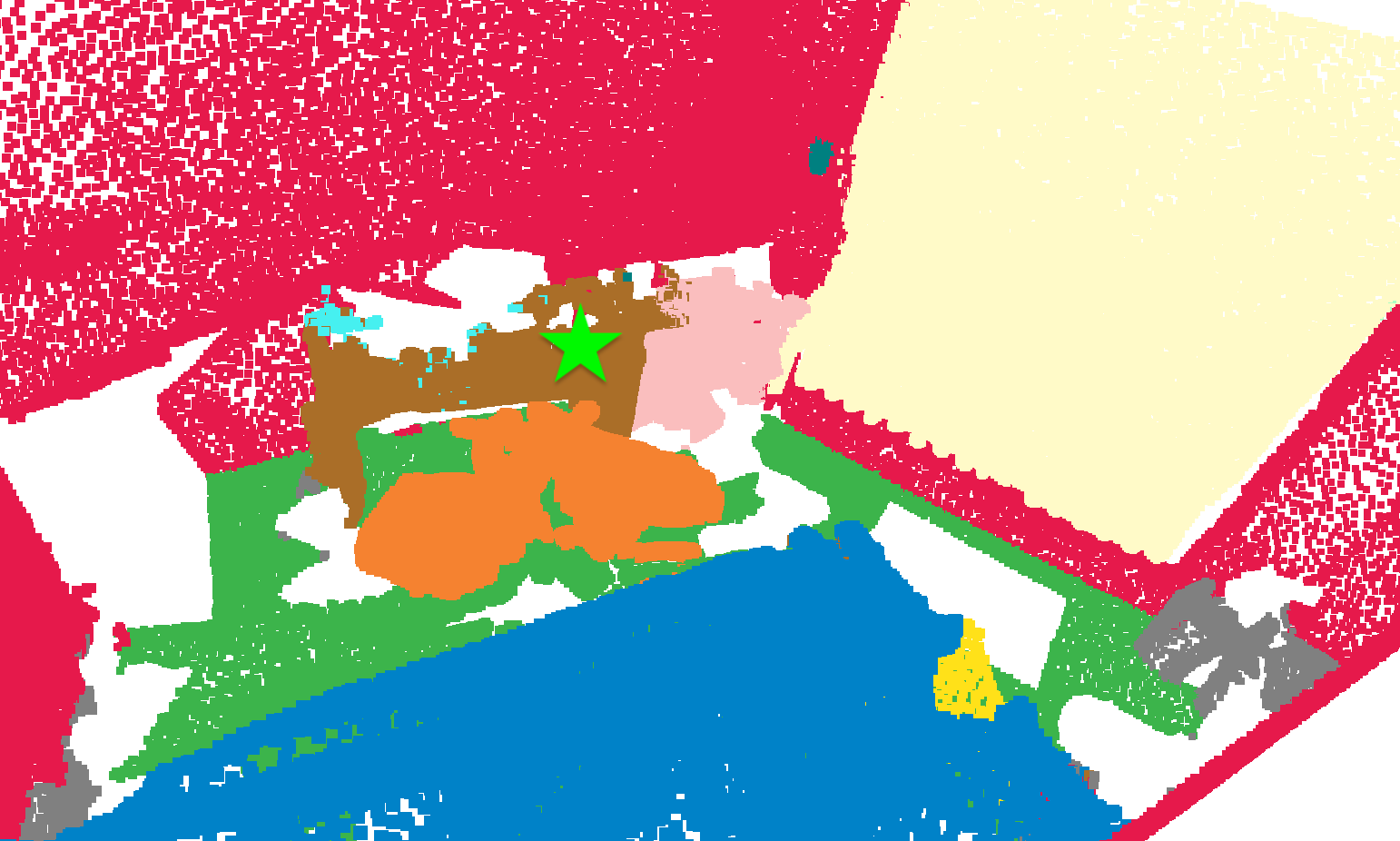}\\\footnotesize w/ stratified
    \end{minipage} 
    
    \vspace{0.1cm}
     
    \begin{minipage}  {0.06\linewidth}
        \centering
        \includegraphics [width=0.5\linewidth,height=0.5\linewidth]
        {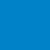}
    \end{minipage}\footnotesize bed
    \begin{minipage}  {0.06\linewidth}
        \centering
        \includegraphics [width=0.5\linewidth,height=0.5\linewidth]
        {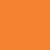}
    \end{minipage}\footnotesize chair
    \begin{minipage}  {0.06\linewidth}
        \centering
        \includegraphics [width=0.5\linewidth,height=0.5\linewidth]
        {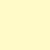}
    \end{minipage}\footnotesize curtain
    \begin{minipage}  {0.06\linewidth}
        \centering
        \includegraphics [width=0.5\linewidth,height=0.5\linewidth]
        {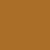}
    \end{minipage}\footnotesize desk
    \begin{minipage}  {0.06\linewidth}
        \centering
        \includegraphics [width=0.5\linewidth,height=0.5\linewidth]
        {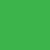}
    \end{minipage}\footnotesize floor
    \begin{minipage}  {0.06\linewidth}
        \centering
        \includegraphics [width=0.5\linewidth,height=0.5\linewidth]
        {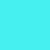}
    \end{minipage}\footnotesize table
    \begin{minipage}  {0.06\linewidth}
        \centering
        \includegraphics [width=0.5\linewidth,height=0.5\linewidth]
        {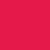}
    \end{minipage}\footnotesize wall
\end{center}
\vspace{-0.5cm}
\caption{Visualization of Effective Receptive Field (ERF)~\cite{luo2016understanding}, given the feature of interest (shown with green star) in the output layer. Red region corresponds to high contribution. \textbf{Left}: Input point cloud and the ground truth. \textbf{Middle}: The ERF and prediction of the model without stratified strategy and by only attending to its own window. \textbf{Right}: The ERF and prediction of the model with direct long-range dependency, using the stratified strategy. More illustrations are shown in the supplementary file.}
\label{fig:erf}
\vspace{-0.4cm}
\end{figure}

Specifically, we first partition the 3D space into non-overlapping cubic windows, inspired by Swin Transformer~\cite{liu2021Swin}. However, in Swin Transformer, different windows work independently, and each query token only chooses the tokens within its window as keys, thus attending to a limited local region. Instead, we propose a stratified strategy for sampling keys. Rather than only selecting nearby points in the same window as keys, we also sparsely sample distant points. In this way, for each query point, both denser nearby points and sparser distant points are sampled to form the keys all together, achieving a significantly enlarged effective receptive field while incurring negligible extra computations. For instance, we visualize the Effective Receptive Field (ERF)~\cite{luo2016understanding} in Fig.~\ref{fig:erf} to show the importance of modeling long-range contexts. In the middle of the figure, due to incapability to model the direct long-range dependency, the \textit{desk} merely attends to the local region, leading to false predictions.
Contrarily, with our proposed stratified strategy, the \textit{desk} is able to aggregate contexts from distant objects, such as the \textit{bed} or \textit{curtain}, which helps to correct the prediction.


Moreover, it is notable that irregular point arrangements pose significant challenges in designing 3D Transformer. In 2D images, patch-wise tokens can be easily formed with spatially regular pixels. But 3D points are completely different. In our framework, each point is deemed as a token and we perform point embedding for each point to aggregate local information in the first layer, which is beneficial for faster convergence and stronger performance. Furthermore, we adopt effective relative position encoding to capture richer position information. It can generate the positional bias dynamically with contexts, through the interaction with the semantic features. Also, 
considering that 3D point numbers in different windows vary a lot and cause unnecessary memory occupation for windows with a small number of points, we introduce a memory-efficient implementation to significantly reduce memory consumption.

In total, our contribution is threefold:
\begin{itemize}

    \vspace{-0.1cm}
    \item We propose Stratified Transformer to additionally sample distant points as keys but in a sparser way, enlarging the effective receptive field and building direct long-range dependency while incurring negligible extra computations.
    
    \vspace{-0.1cm}
    \item To handle irregular point arrangements, we design first-layer point embedding and effective contextual position encoding, along with a memory-efficient implementation, to build a strong Transformer tailored for 3D point cloud segmentation. 
    
    \vspace{-0.1cm}
    \item Experiments show our model achieves state-of-the-art results on widely adopted large-scale segmentation datasets, \ie, S3DIS~\cite{armeni_cvpr16}, ScanNetv2~\cite{dai2017scannet} and ShapeNetPart~\cite{chang2015shapenet}. Extensive ablation studies verify the benefit of each component.
\end{itemize}

\section{Related Work}


\paragraph{Vision Transformer.} Recently, vision Transformer~\cite{vaswani2017attention} becomes popular in 2D image understanding~\cite{dosovitskiy2020vit, pmlr-v139-touvron21a, touvron2021cait, wang2021pyramid, wang2021pvtv2, liu2021Swin, chu2021Twins, yang2021focal, dong2021cswin, vip, detr, zhu2020deformable, mao2021voxel, zhao2020san}. ViT~\cite{dosovitskiy2020vit} treats each patch as a token, and directly uses a Transformer encoder to extract features for image classification. Further, PVT~\cite{wang2021pyramid} proposes a hierarchical structure to obtain a pyramid of features for semantic segmentation and also presents Spatial Reduction Attention to save memory. Alternatively, Swin Transformer~\cite{liu2021Swin} uses a window-based attention, and proposes a shifted window operation in the successive Transformer block. Methods of~\cite{chu2021Twins, yang2021focal, dong2021cswin} further propose different designs to incorporate long-range and global dependencies. Transformer is already popular in 2D, but remains under-explored on point clouds. Inspired by Swin Transformer, we adopt hierarchical structure and shifted window operation for 3D point cloud. On top of that, we propose a stratified strategy for sampling keys to harvest long-range contexts, and put forward several essential designs to combat the challenges posed by irregular point arrangements. 

\vspace{-0.4cm}
\paragraph{Point Cloud Segmentation.} Approaches for point cloud segmentation can be grouped into two categories, \ie, voxel-based and the point-based methods. The voxel-based solutions~\cite{3DSemanticSegmentationWithSubmanifoldSparseConvNet, SubmanifoldSparseConvNet, choy20194d} first divide the 3D space into regular voxels, and then apply sparse convolutions upon them. 
They yield decent performance, but suffer from inaccurate position information due to voxelization.
Point-based methods~\cite{qi2017pointnet, qi2017pointnet++, wu2019pointconv, thomas2019kpconv, zhao2020point, zhao2019pointweb, dai20183dmv, narita2019panopticfusion, li2018pointcnn, chiang2019unified, yan2020pointasnl, lei2020seggcn, hu2020randla, hu2020jsenet, zhang2020deep, tchapmi2017segcloud, tatarchenko2018tangent, jiang2019hierarchical, wang2019graph, yang2019modeling, wang2018deep, landrieu2018large, xu2021paconv, atzmon2018point, xu2018spidercnn, wang2019dynamic, liu2019point2sequence, mao2019interpolated, liu2019densepoint, su2018splatnet, liu2021group, engelmann2020dilated, chu2021icm} directly adopt the point features and positions as inputs, thus keeping the position information intact.
Following this line of research, different ways for feature aggregation are designed to learn high-level semantic features.
PointNet and its variants~\cite{qi2017pointnet, qi2017pointnet++} use max pooling to aggregate features. PointConv~\cite{wu2019pointconv} and KPConv~\cite{thomas2019kpconv} try to use an MLP or discrete kernel points to mimic a continuous convolution kernel. Point Transformer~\cite{zhao2020point} uses the ``vector self-attention" operator to aggregate local features and the ``subtraction relation" to generate the attention weights, but it suffers from lack of long-range contexts and insufficient robustness upon various perturbations in testing. 

Our work is pointed-based and closely related point transformer yet with a fundamental difference: ours overcomes the limited effective receptive field issue and makes the best of Transformer for modeling long-range contextual dependencies instead of merely local aggregation.

\begin{figure*}
\begin{center}
\includegraphics[width=1.0\linewidth]{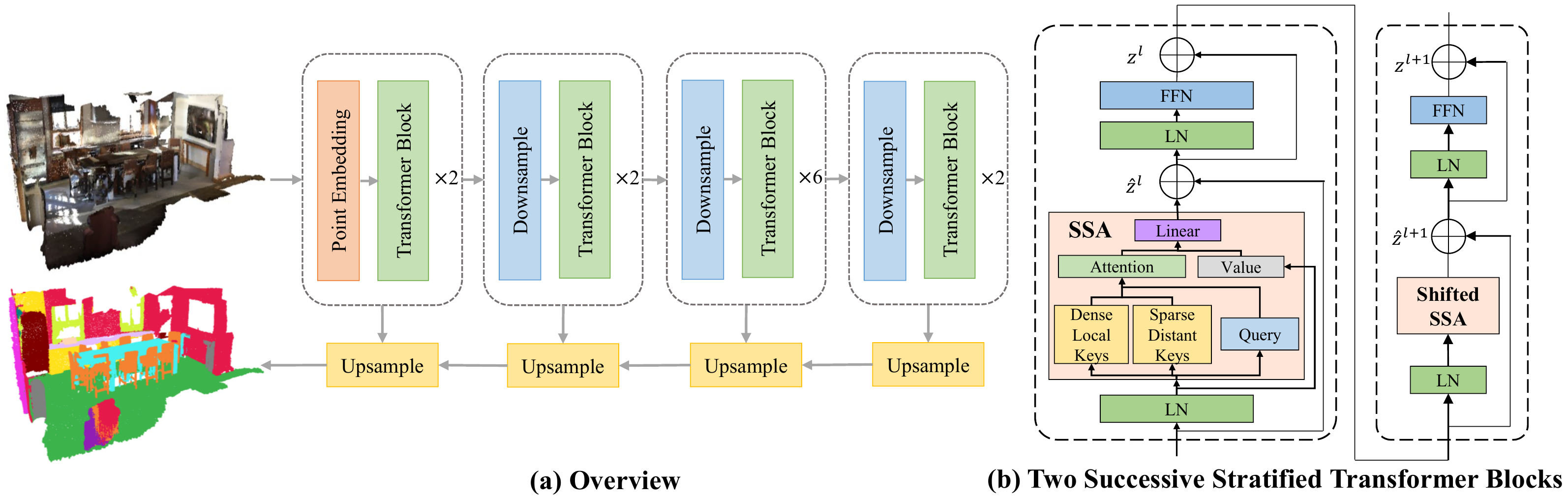}
\end{center}
\vspace{-0.5cm}
\caption{(a) Framework Overview. (b) Structure of Stratified Transformer Block. Hierarchical structure is employed to obtain multi-level features. Input point clouds firstly go through the Point Embedding module to aggregate local structure information. After several downsample layers and transformer blocks, the features are upsampled for segmentation. SSA: Stratified Self-attention. Shifted SSA: SSA with shifted window. Best viewed in color.}
\label{fig:overview}
\vspace{-0.3cm}
\end{figure*}

\section{Our Method}



\subsection{Overview}
\label{sec:overview}

The overview of our model is illustrated in Fig.~\ref{fig:overview}. Our framework is point-based, and we use both \textit{xyz} coordinates and \textit{rgb} colors as input. The encoder-decoder structure is adopted where the encoder is composed of multiple stages connected by downsample layers. At the beginning of the encoder, the first-layer point embedding module is used for local aggregation. Then, there are several Transformer blocks at each stage. As for the decoder, the encoder features are upsampled to become denser layer by layer in the way similar to U-Net~\cite{ronneberger2015u}. 





\subsection{Transformer Block}
\label{sec:transformer_block}
The Transformer block is composed of a standard multi-head self-attention module and a feed-forward network (FFN). With tens of thousands of points as inputs, directly applying global self-attention incurs unacceptable $O(N^2)$ memory consumption, where $N$ is the input point number.

\vspace{-0.3cm}
\paragraph{Vanilla Version.}
To this end, we employ window-based self-attention. The 3D space is firstly partitioned into non-overlapping cubic windows, where the points are scattered in different windows. Instead of attending to all the points as in global self-attention, each query point only needs to consider neighbors in the same window. Multi-head self-attention is performed in each window independently. Since different windows may contain varying numbers of points, we denote $k_t$ as the number of points within the $t$-th window. Formally, given that $N_h$ is the number of heads, $N_d$ is the dimension of each head and $N_c=N_h \times N_d$ is the feature dimension, for the input points in the $t$-th window $\mathbf{x} \in \mathbb{R}^{k_t\times (N_h \times N_d)}$, the multi-head self-attention in the $t$-th window is formulated as
\begin{equation}
\footnotesize
\label{eq_sa_first}
    \mathbf{q} = Linear_q(\mathbf{x}), \quad \mathbf{k} = Linear_k(\mathbf{x}), \quad \mathbf{v} = Linear_v(\mathbf{x}),\nonumber
\end{equation}
\vspace{-0.8cm}
\begin{eqnarray}
\footnotesize
\label{eq_sa_attn}
    \mathbf{attn}_{i,j,h} &=& \mathbf{q}_{i,h} \cdot \mathbf{k}_{j,h},\nonumber\\
    \hat{\mathbf{attn}}_{i,.,h} &=& softmax(\mathbf{attn}_{i,.,h}),\\
\label{eq_sa_last}
\vspace{-0.2cm}
    \mathbf{y}_{i,h} &=& \sum_{j=1}^{k_t} \hat{\mathbf{attn}}_{i,j,h} \times \mathbf{v}_{j,h},\nonumber
\end{eqnarray}
\vspace{-0.2cm}
\begin{equation}
\footnotesize
    \hat{\mathbf{z}} = Linear(\mathbf{y}),\nonumber
\end{equation}
where $\mathbf{q}, \mathbf{k}, \mathbf{v} \in \mathbb{R}^{k_t\times N_h \times N_d}$ are obtained from $\mathbf{x}$ by three linear layers, and $\cdot$ means dot product between vectors $\mathbf{q}_{i,h}$ and $\mathbf{k}_{j,h}$. $\mathbf{attn} \in \mathbb{R}^{k_t\times k_t \times N_h}$ is the attention map, and $\mathbf{y}\in \mathbb{R}^{k_t\times N_h \times N_d}$ is the aggregated feature, which is further projected to the output feature $\hat{\mathbf{z}}\in \mathbb{R}^{k_t\times (N_h \times N_d)}$. 

Note that the above equations only show the calculation in a single window, and different windows work in the same way independently. In this way, the memory complexity is dramatically reduced to $O(\frac{N}{k}\times k^2)=O(N\times k)$, where $k$ is the average number of points scattered in each window.

To facilitate cross-window communication, we also shift the window by half of the window size between two successive Transformer blocks, similar to \cite{liu2021Swin}. The illustration of shifted window is given in the supplementary file.

\begin{figure}
\begin{center}
\includegraphics[width=0.95\linewidth]{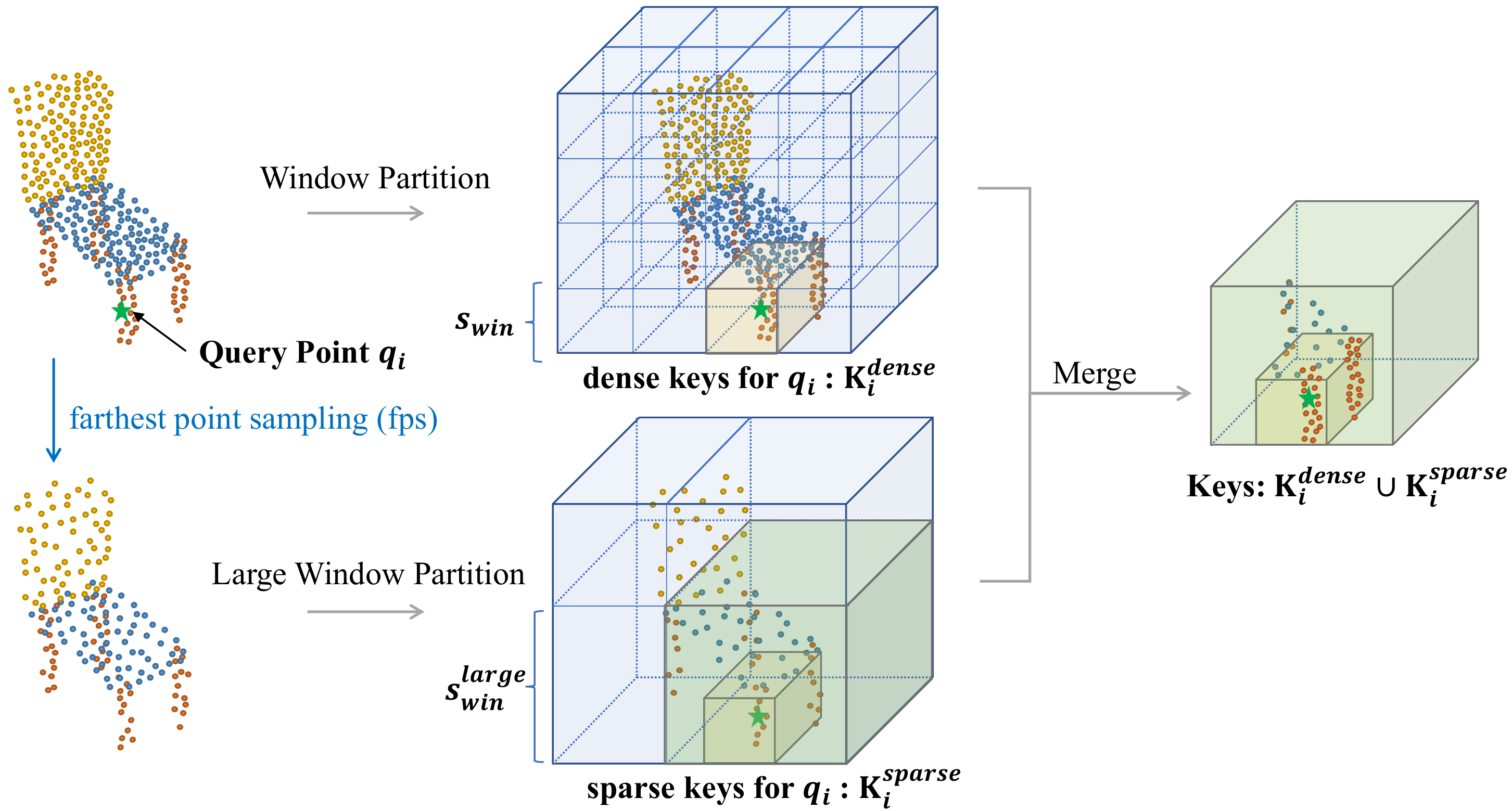}
\end{center}
\vspace{-0.5cm}
\caption{Illustration of the stratified strategy for keys sampling. The green star denotes the given query point.}
\label{fig:stratified}
\vspace{-0.4cm}
\end{figure}

\vspace{-0.5cm}
\paragraph{Stratified Key-sampling Strategy.}
Since every query point only attends to the local points in its own window, the vanilla version Transformer block suffers from limited effective receptive field even with shifted window, as shown in Fig.~\ref{fig:erf}. Therefore, it fails to capture long-range contextual dependencies over distant objects, causing false predictions.

A simple solution is to enlarge the size of cubic window. However, the memory would grow as the window size increases. To effectively aggregate long-range contexts at a low cost of memory, we propose a stratified strategy for sampling keys.
As shown in Fig.~\ref{fig:stratified}, we partition the space into non-overlapping cubic windows with the window size $s_{win}$. For each query point $q_i$ (shown with green star), we find the points $\mathbf{K}_{i}^{dense}$ in its window, same as the vanilla version. Additionally, we downsample the input points through farthest point sampling (fps) at the scale of $s$, and find the points $\mathbf{K}_{i}^{sparse}$ with a larger window size $s_{win}^{large}$. In the end, both dense and sparse keys form the final keys, \ie, $\mathbf{K}_{i} = \mathbf{K}_{i}^{dense} \cup \mathbf{K}_{i}^{sparse}$. Note that duplicated key points are only counted once.


The complete structure of Stratified Transformer block is shown in Fig.~\ref{fig:overview} (b). Following common practice, we use LayerNorm~\cite{ba2016layer} before each self-attention module or feed-forward network. To further complement the information interaction across windows, the original window is shifted by $\frac{1}{2}s_{win}$ while the large window is shifted by $\frac{1}{2}s_{win}^{large}$ in the successive Transformer block. This further boosts the performance as listed in Table~\ref{table:shifted_window}.

Thanks to the stratified strategy for key sampling, the effective receptive field is enlarged remarkably and the query feature is able to effectively aggregate long-range contexts. Compared to the vanilla version, we merely incur the extra computations on the sparse distant keys, which only takes up about 10\% of the final keys $\mathbf{K}_{i}$.


\subsection{First-layer Point Embedding}
\label{sec:point_emb}

\begin{figure}
\begin{center}
\includegraphics[width=0.9\linewidth,height=0.45\linewidth]{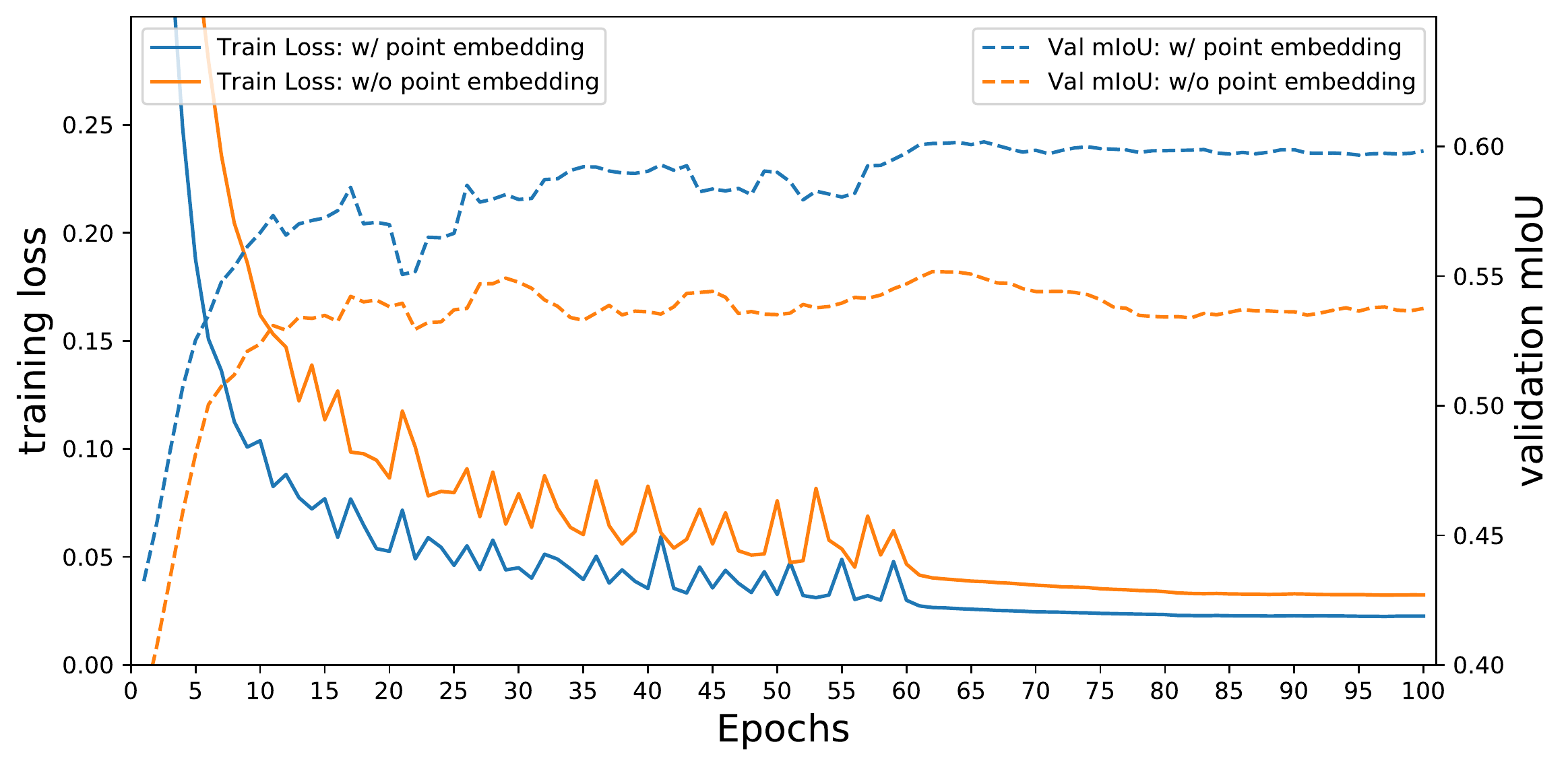}
\end{center}
\vspace{-0.7cm}
\caption{Plot of training loss (solid line) and validation mIoU (dotted line) in the training process. The models w/ (blue curve) and w/o (orange curve) first-layer point embedding are compared. 
}
\label{fig:point_emb_plot}
\vspace{-0.4cm}
\end{figure}

In the first layer, we build a point embedding module. An intuitive choice is to use a linear layer or MLP to project the input features to a high dimension. However, we empirically observe relatively slow convergence and poor performance by using a linear layer in the first layer, as shown in Fig.~\ref{fig:point_emb_plot}. We note that the point feature from a linear layer or MLP merely comprises the raw information of its own \textit{xyz} position and the \textit{rgb} color, but it lacks local geometric and contextual information. As a result, in the first Transformer block, the attention map could not capture high-level relevance between the queries and keys that only contain raw \textit{xyz} and \textit{rgb} information. This negatively affects representation power and generalization ability of the model.




We contrarily propose to aggregate the features of local neighbors for each point in the Point Embedding module. 
We try a variety of methods for local aggregation, such as max pooling and average pooling, and find KPConv performs the best, as shown in Table~\ref{table:exp_local_agg}.
Surprisingly, this minor modification to the architecture brings about considerable improvement as suggested in Exp.\uppercase\expandafter{\romannumeral1} and \uppercase\expandafter{\romannumeral2} as well as Exp.\uppercase\expandafter{\romannumeral5} and \uppercase\expandafter{\romannumeral6} of Table~\ref{tab:ablation}. It proves the importance of initial local aggregation in the Transformer-based networks. Note that a single KPConv incurs negligible extra computations (merely $2\%$ FLOPs) compared to the whole network.


\subsection{Contextual Relative Position Encoding}
\label{sec:crpe}
Compared to 2D spatially regular pixels, 3D points are in a more complicated continuous space, posing challenges to exploit the \textit{xyz} position. \cite{3detr} claims that position encoding is unnecessary for 3D Transformer-based networks because the \textit{xyz} coordinates have already been used as the input features. However, although the input of the Transformer block has already contained the \textit{xyz} position, fine-grained position information may be lost in high-level features when going deeper through the network. To make better use of the position information, we adopt a context-based adaptive relative position encoding scheme inspired by \cite{wu2021rethinking}. 


Particularly, for the point features $\mathbf{x} \in \mathbb{R}^{k_t\times (N_h \times N_d)}$ in the $t$-th window, we denote the \textit{xyz} coordinates as $\mathbf{p} \in \mathbb{R}^{k_t \times 3}$. So, the relative $xyz$ coordinates $\mathbf{r} \in \mathbb{R}^{k_t \times k_t \times 3}$ between the queries and keys are formulated as 

\vspace{-0.3cm}
\begin{footnotesize}
\begin{equation}
    \mathbf{r}_{i,j,m} = \mathbf{p}_{i,m} - \mathbf{p}_{j,m}, \quad 1 \le i,j \le k_t, m \in \{1,2,3\}.
\end{equation}
\end{footnotesize}
\vspace{-0.3cm}

To map relative coordinates to the corresponding position encoding, we maintain three learnable look-up tables $\mathbf{t}_x, \mathbf{t}_y, \mathbf{t}_z \in \mathbb{R}^{L\times (N_h\times N_d)}$ corresponding to $x$, $y$ and $z$ axis, respectively. As the relative coordinates are continuous floating-point numbers, we uniformly quantize the range of $\mathbf{r}_{i,j,m}$, \ie, $(-s_{win}, s_{win})$ into $L$ discrete parts and map the relative coordinates $\mathbf{r}_{i,j,m}$ to the indices of the tables as

\vspace{-0.25cm}
\begin{equation}
\footnotesize
    \mathbf{idx}_{i,j,m} = \lfloor \frac{\mathbf{r}_{i,j,m} + s_{win}}{s_{quant}} \rfloor,
\end{equation}
where $s_{win}$ is the window size and $s_{quant}=\frac{2\cdot s_{win}}{L}$ is the quantization size, and $\lfloor \cdot \rfloor$ denotes floor rounding. 

We look up the tables to retrieve corresponding embedding with the index and sum them up to obtain the position encoding of

\vspace{-0.15cm}
\begin{equation}
\footnotesize
    \mathbf{e}_{i,j} = \mathbf{t}_x[\mathbf{idx}_{i,j,1}] + \mathbf{t}_y[\mathbf{idx}_{i,j,2}] + \mathbf{t}_z[\mathbf{idx}_{i,j,3}],
\vspace{0.1cm}
\end{equation}
where $\mathbf{t}[idx] \in \mathbb{R}^{N_h\times N_d}$ means the $idx$-th entry of the table $\mathbf{t}$, and $\mathbf{e} \in \mathbb{R}^{k_t \times k_t \times N_h \times N_d}$ is the position encoding.

Practically, the tables for query, key and value are not shared. So we differentiate among them by adding a superscript, where $\mathbf{t}_x^q$ denotes the $x$-axis table for the query. Similarly, the position encoding corresponding to query, key and value is denoted by $\mathbf{e}^q$, $\mathbf{e}^k$ and $\mathbf{e}^v$, respectively.

Then the position encoding performs dot product with the query and key feature to obtain the positional bias $\mathbf{pos\_bias} \in \mathbb{R}^{k_t\times k_t \times N_h}$, which is then added to the attention map. Also, we add the value feature with its corresponding position encoding, followed by the weighted sum aggregation.
Finally, the original equations Eq.~\eqref{eq_sa_attn} are updated to the contextual Relative Position Encoding (cRPE) version of

\vspace{-0.5cm}
\begin{footnotesize}
\begin{align}
    \mathbf{pos\_bias}_{i,j,h}^{cRPE} &= \mathbf{q}_{i,h} \cdot \mathbf{e}_{i,j,h}^q + \mathbf{k}_{j,h} \cdot \mathbf{e}_{i,j,h}^k,\nonumber\\
    \mathbf{attn}_{i,j,h}^{cRPE} &= \mathbf{q}_{i,h} \cdot \mathbf{k}_{j,h} + \mathbf{pos\_bias}_{i,j,h}^{cRPE},\nonumber\\
    \hat{\mathbf{attn}}_{i,.,h}^{cRPE} &= softmax(\mathbf{attn}_{i,.,h}^{cRPE}),\nonumber\\
    \mathbf{y}_{i,h}^{cRPE} &= \sum_{j=1}^{k_t} \hat{\mathbf{attn}}_{i,j,h}^{cRPE} \times (\mathbf{v}_{j,h} + \mathbf{e}_{i,j,h}^v).\nonumber
\end{align}
\end{footnotesize}
\vspace{-0.2cm}


\begin{figure}
\begin{center}
	\centering
    \begin{minipage}  {0.28\linewidth}
        \centering
        \includegraphics [width=1\linewidth,height=0.72\linewidth]
        {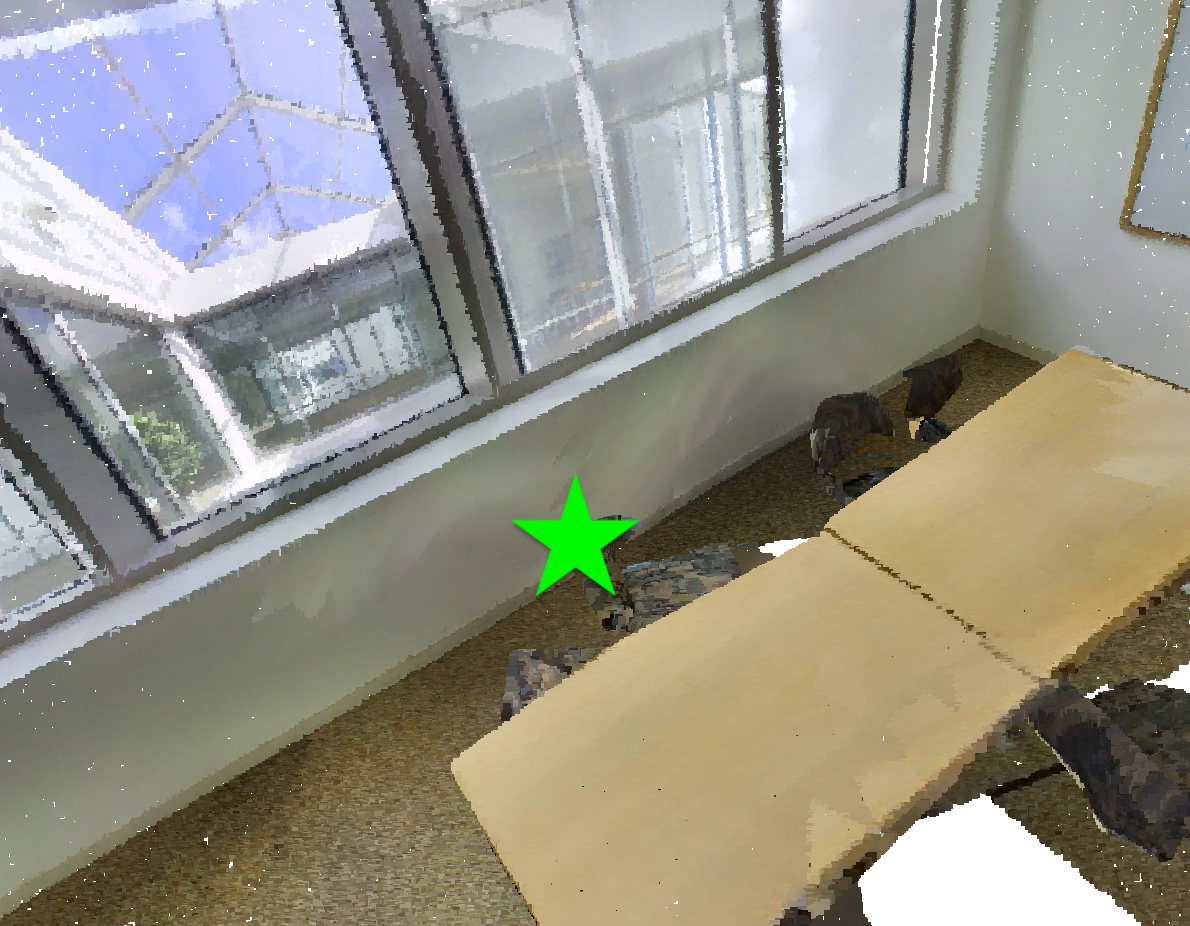}\\\footnotesize Input
    \end{minipage}      
    \begin{minipage}  {0.28\linewidth}
        \centering
        \includegraphics [width=1\linewidth,height=0.72\linewidth]
        {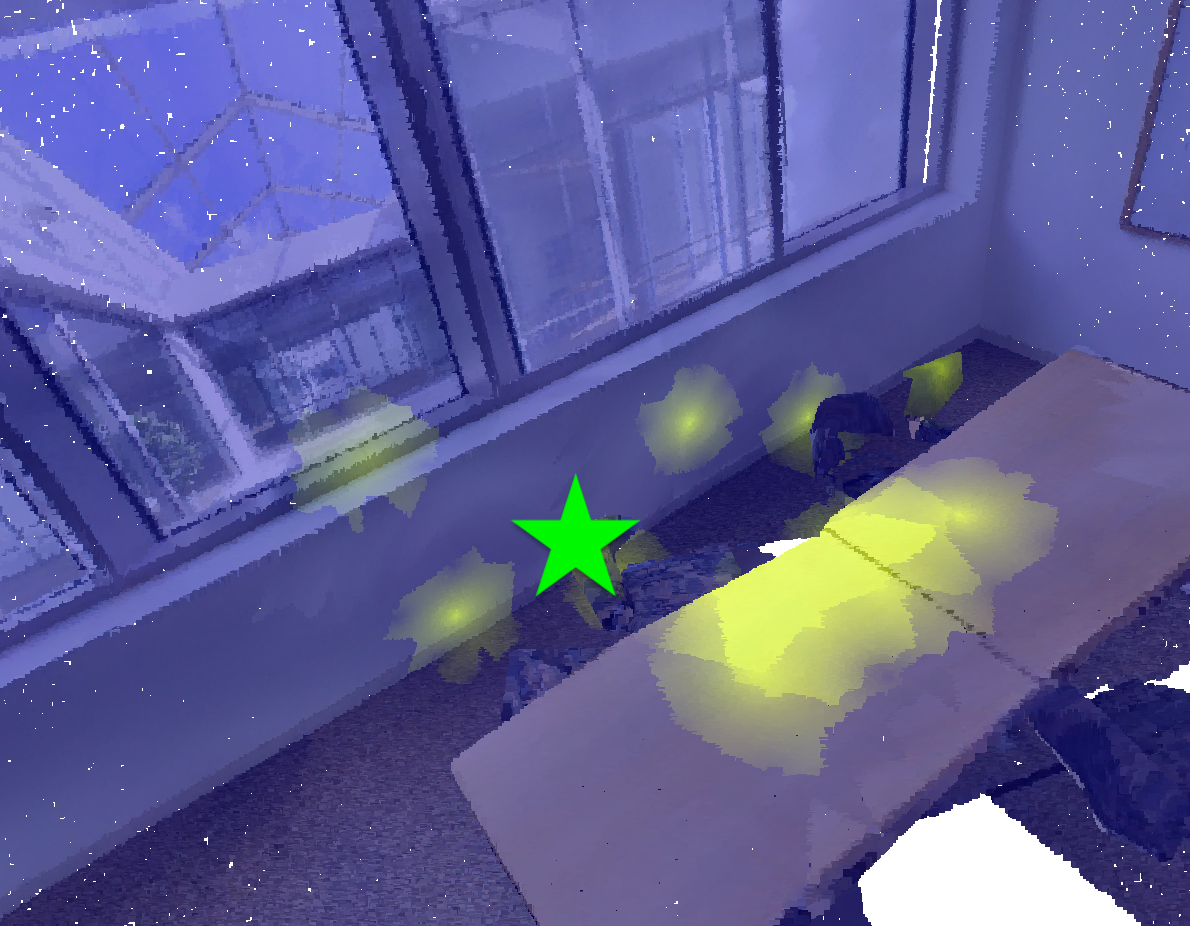}\\\footnotesize MLP-based
    \end{minipage}      
     \begin{minipage}  {0.28\linewidth}
        \centering
        \includegraphics [width=1\linewidth,height=0.72\linewidth]
        {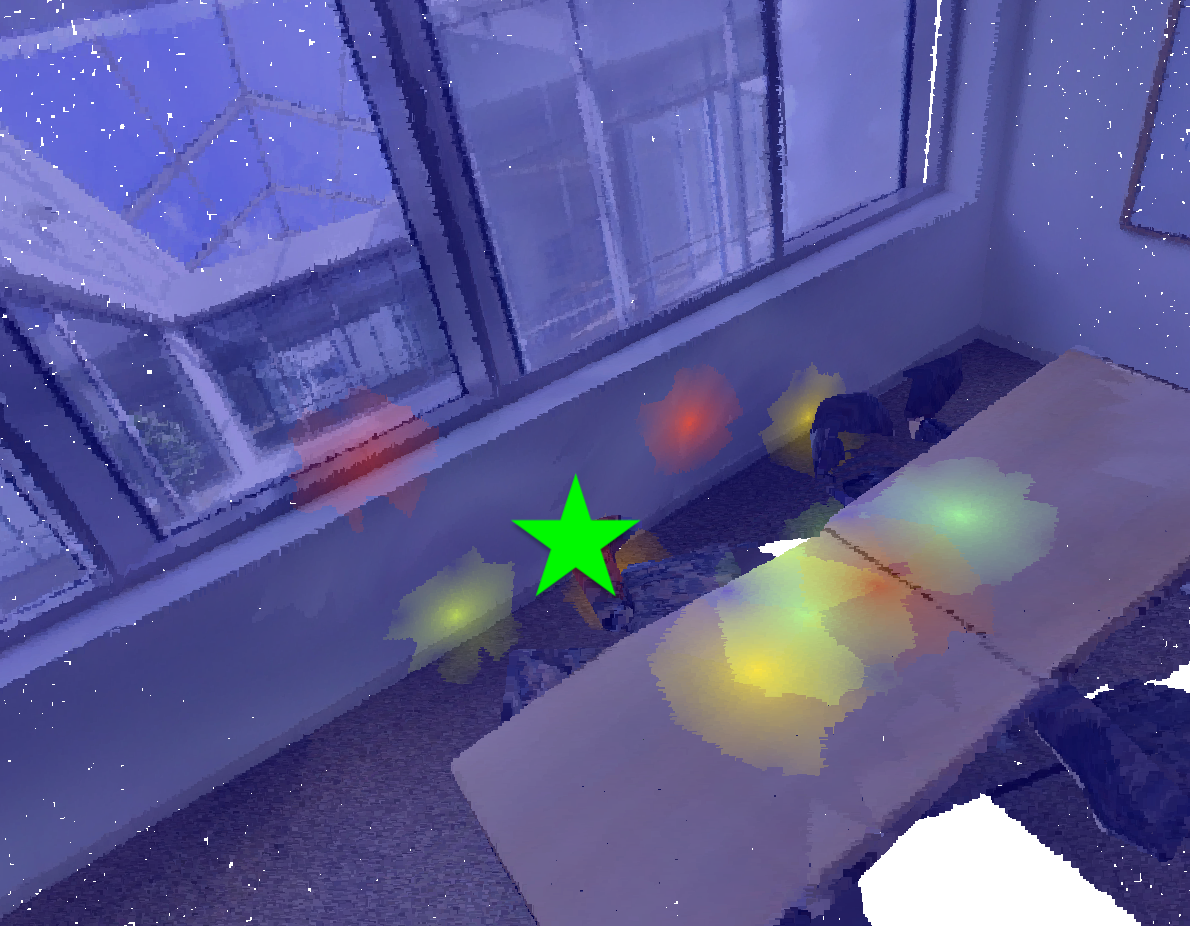}\\\footnotesize cRPE
    \end{minipage}
     \begin{minipage}  {0.13\linewidth}
        \centering
        \includegraphics [width=0.4\linewidth,height=1.6\linewidth]
        {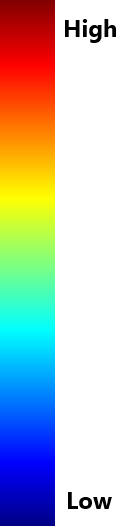}
    \end{minipage}
    
\end{center}
\vspace{-0.4cm}
\caption{Visualization of the positional bias of each key at the first head of the last transformer block given the query point (shown with green star). The color map is shown on the right.}
\label{fig:rpe}
\vspace{-0.3cm}
\end{figure}

Compared to the MLP-based position encoding\label{sec:crpe}, where the relative \textit{xyz} coordinates $\mathbf{r} \in \mathbb{R}^{k_t\times k_t \times 3}$ are directly projected to the positional bias $\mathbf{pe\_bias} \in \mathbb{R}^{k_t\times k_t \times N_h}$ via an MLP, cRPE adaptively generates the positional bias through the dot product with queries and keys, thus providing semantic information. 
The positional bias of the MLP-based and cRPE are visualized in Fig.~\ref{fig:rpe}. It reveals the fact that the positional bias generated by the MLP-based model is similar among the keys. So it makes little difference to the attention weights. But for cRPE, the positional bias varies a lot for different keys. Besides, Exp.~\uppercase\expandafter{\romannumeral3} and \uppercase\expandafter{\romannumeral4} and Exp.~\uppercase\expandafter{\romannumeral5} and \uppercase\expandafter{\romannumeral8} of Table~\ref{tab:ablation} also show the superiority of cRPE.

\begin{figure}
\begin{center}
\includegraphics[width=0.95\linewidth]{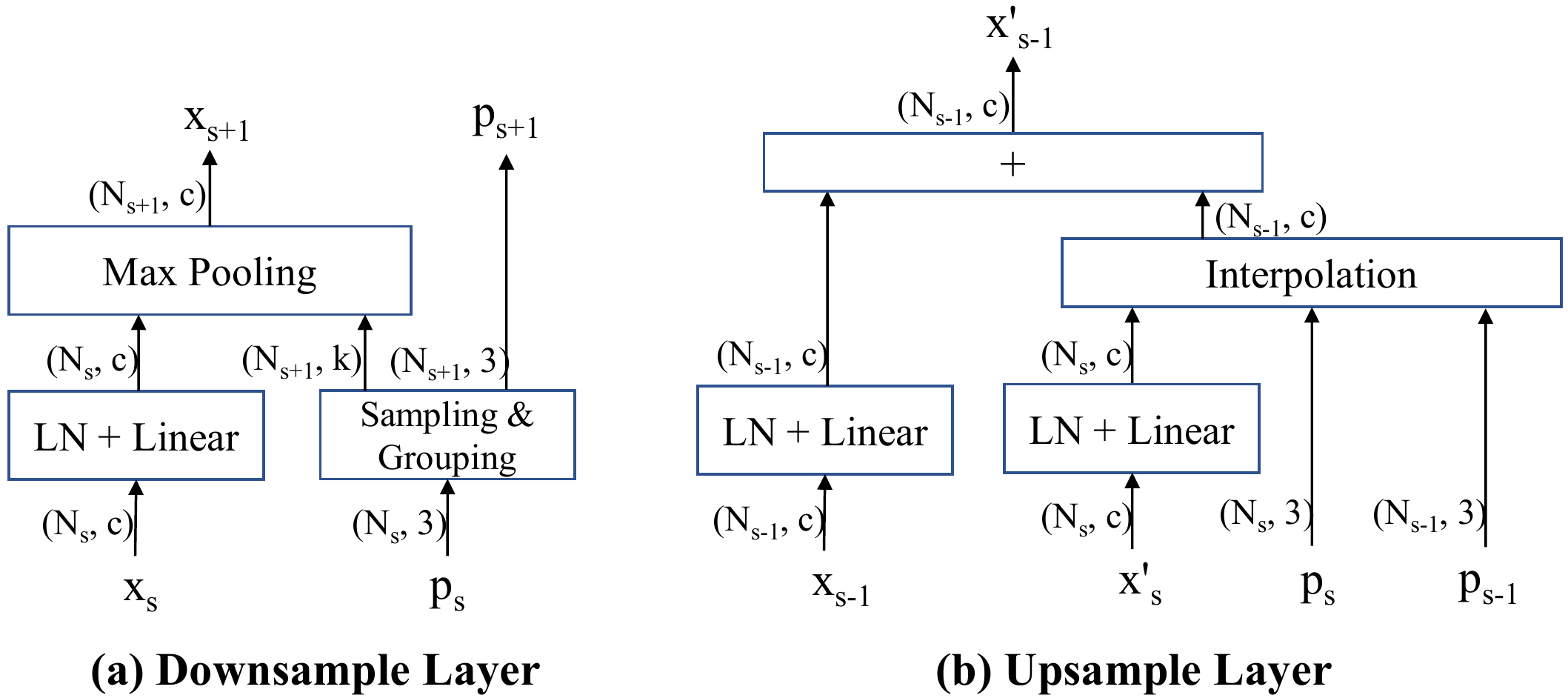}
\end{center}
\vspace{-0.5cm}
\caption{Structural illustration of (a) Downsample Layer and (b) Upsample Layer.}
\label{fig:down_up}
\vspace{-0.3cm}
\end{figure}

\subsection{Downsample and Upsample Layers}
\label{sec:down_up}
The Downsample Layer is shown in Fig.~\ref{fig:down_up} (a). First, the \textit{xyz} coordinates $\mathbf{p}_s$ go through the Sampling \& Grouping module, where we first sample centroid points $\mathbf{p}_{s+1}$ by fathest point sampling (fps)~\cite{qi2017pointnet++} and then use kNN to query the original points to get the grouping index $\mathbf{idx_{group}} \in \mathbb{R}^{N_{s+1}\times k}$. The number of centroid points is $\frac{1}{4}$ of the original points, \ie, $N_{s+1} = \lceil \frac{1}{4} N_s \rceil$. Meanwhile, the point features $\mathbf{x}_s$ are fed into a Pre-LN~\cite{xiong2020layer} linear projection layer. Further, we exploit max pooling to aggregate the projected features using the grouping index, yielding the output features $\mathbf{x}_{s+1}$.


For the upsample layer, as shown in Fig.~\ref{fig:down_up} (b), the decoder features $\mathbf{x}'_s$ are firstly projected by a Pre-LN linear layer. We perform interpolation~\cite{qi2017pointnet++} between current \textit{xyz} coordinates $\mathbf{p}_s$ and the previous ones $\mathbf{p}_{s-1}$. The encoder point features in the previous stage $\mathbf{x}_{s-1}$ go through a Pre-LN linear layer. Finally, we sum them up to yield the next decoder features $\mathbf{x}'_{s-1}$.

\section{Memory-efficient Implementation}
\label{sec:imp}

\begin{figure}
\begin{center}
\includegraphics[width=1.0\linewidth]{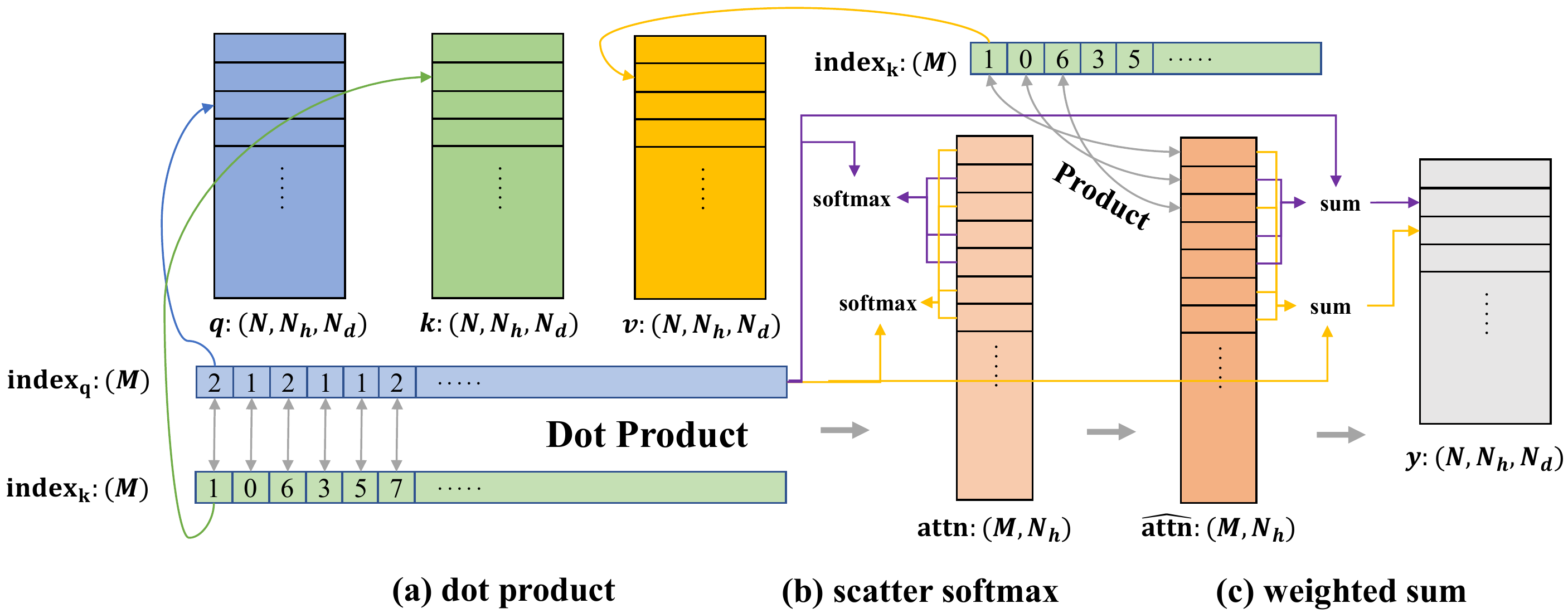}
\end{center}
\vspace{-0.6cm}
\caption{Memory-efficient implementation includes three steps: (a) dot product; (b) scatter softmax; (c) weighted sum. It is best viewed in color and by zoom-in.}
\label{fig:imp}
\vspace{-0.4cm}
\end{figure}

In 2D Swin Transformer, it is easy to implement the window-based attention because the number of tokens is fixed in each window. Nevertheless, due to the irregular point arrangements in 3D, the number of the tokens in each window varies a lot. A simple solution is to pad the tokens in each window to the maximum token number $k_{max}$ with dummy tokens, and then apply a masked self-attention. But this solution wastes much memory and computations.

Instead, we first pre-compute all pairs of query and key that need to perform dot product. As shown in Fig.~\ref{fig:imp} (a), we use two indices of $\mathbf{index}_q, \mathbf{index}_k \in \mathbb{R}^{M}$, to index the $\mathbf{q}$ and $\mathbf{k}$ of shape $(N, N_h, N_d)$, respectively, where $N$ denotes the total number of input points. Then, we perform dot product between the entries indexed by $\mathbf{index}_q$ and $\mathbf{index}_k$, yielding the attention map $\mathbf{attn}$ of the shape $(M, N_h)$.
Afterwards, as shown in Fig.~\ref{fig:imp} (b), we perform the scatter softmax directly on $\mathbf{attn}$ with the query index $\mathbf{index}_q$, where the softmax function is applied on the entries in $\mathbf{attn}$ with the same index in $\mathbf{index}_q$. Further, as shown in Fig.~\ref{fig:imp} (c), we use $\mathbf{index}_k$ to index the values $\mathbf{v}$ and multiply them with the attention map $\mathbf{attn}$. We finally sum up the entries with the same index in $\mathbf{index}_q$ and save the results into the output features $\mathbf{y}$. Note that each of the steps is implemented by a single CUDA kernel. So the intermediate variables inside each step hardly occupy memory. In this way, we reach the memory complexity of $O(M \cdot N_h)$, much less than that used in vanilla implementation. More detailed memory complexity analysis and discussion of position encoding implementation are given in the supplementary file. Our implementation saves 57\% memory compared to the vanilla one.

\section{Experiments}

\subsection{Experimental Setting}
\label{exp:setting}

\paragraph{Network Architecture.} The main architecture is shown in Fig.~\ref{fig:overview}. Both the \textit{xyz} coordinates and \textit{rgb} colors are used as inputs. We set the initial feature dimension and number of heads to 48 and 3 respectively, and they will double in each downsample layer. As for S3DIS, four stages are constructed with the block depths [2, 2, 6, 2]. In contrast, for ScanNetv2, we note that the point number is larger. So we add an extra downsample layer on top of the first-layer point embedding module. Then, the later four stages with block depths [3, 9, 3, 3] are added. So a total of five stages are constructed for ScanNetv2. 

\vspace{-0.3cm}
\paragraph{Implementation Detail.} 
For S3DIS, following previous work~\cite{zhao2020point}, we train for $76,500$ iterations with 4 RTX 2080Ti GPUs. The batch size is set to $8$. Following common practice, the raw input points are firstly grid sampled with the grid size set to $0.04$m. During training, the maximum input points number is set to $80,000$, and all extra ones are discarded if points number reaches this number. The window size is set to $0.16$m initially, and it doubles after each downsample layer. The downsample scale for the stratified sampling strategy is set to $8$. Unless otherwise specified, we use z-axis rotation, scale, jitter and drop color as data augmentation. 

For ScanNetv2, we train for $600$ epochs with weight decay and batch size set to $0.1$ and $8$ respectively, and the grid size for grid sampling is set to $0.02$m. At most $120,000$ points of a point cloud are fed into the network during training. The initial window size is set to $0.1$m. And the downsample scale for the stratified sampling is set to $4$. Except random jitter, the data augmentation is the same as that on S3DIS. The implementation details for ShapeNetPart and the datasets descriptions are given in the supplementary file.


\begin{table}
\small
\begin{center}
\begin{tabular}{l | c | c  c  c }
\toprule 


Method & Input & OA & mAcc & mIoU \\

\specialrule{0em}{2pt}{0pt}
\hline
\specialrule{0em}{2pt}{0pt}
PointNet~\cite{qi2017pointnet} & point & - & 49.0 & 41.1\\

SegCloud~\cite{tchapmi2017segcloud} & point & - & 57.4 & 48.9 \\

TangentConv~\cite{tatarchenko2018tangent} & point & - & 62.2 & 52.6 \\

PointCNN~\cite{li2018pointcnn} & point & 85.9 & 63.9 & 57.3\\

PointWeb~\cite{zhao2019pointweb} & point & 87.0 & 66.6 & 60.3\\

HPEIN~\cite{jiang2019hierarchical} & point & 87.2 & 68.3 & 61.9 \\

GACNet~\cite{wang2019graph} & point & 87.8 & - & 62.9 \\

PAT~\cite{yang2019modeling} & point & - & 70.8 & 60.1 \\

ParamConv~\cite{wang2018deep} & point & - & 67.0 & 58.3 \\

SPGraph~\cite{landrieu2018large} & point & 86.4 & 66.5 & 58.0 \\

SegGCN~\cite{lei2020seggcn} & point & 88.2 & 70.4 & 63.6 \\

MinkowskiNet~\cite{choy20194d} & voxel & - & 71.7 & 65.4 \\

PAConv~\cite{xu2021paconv} & point & - & - & 66.6 \\

KPConv~\cite{thomas2019kpconv} & point & - & 72.8 & 67.1 \\

PointTransformer~\cite{zhao2020point} & point & 90.8 & 76.5 & 70.4 \\

\specialrule{0em}{2pt}{0pt}
\hline
\specialrule{0em}{2pt}{0pt}

Ours & point & \textbf{91.5} & \textbf{78.1} & \textbf{72.0} \\


\bottomrule                    

\end{tabular}
\end{center}
\vspace{-0.5cm}
\caption{Results on S3DIS Area5 for semantic segmentation.}
\label{table:exp_s3dis}
\vspace{-0.2cm}
\end{table}

\begin{table}
\small
\begin{center}
\begin{tabular}{l | c | c c }
\hline
\toprule Method & Input & Val mIoU & Test mIoU \\
\specialrule{0em}{2pt}{0pt}
\hline
\specialrule{0em}{2pt}{0pt}

PointNet++~\cite{qi2017pointnet++} & point & 53.5 & 55.7\\

3DMV~\cite{dai20183dmv} & point & - & 48.4 \\

PanopticFusion~\cite{narita2019panopticfusion} & point & - & 52.9 \\

PointCNN~\cite{li2018pointcnn} & point & - & 45.8 \\

PointConv~\cite{wu2019pointconv} & point & 61.0 & 66.6 \\

JointPointBased~\cite{chiang2019unified} & point & 69.2 & 63.4 \\

PointASNL~\cite{yan2020pointasnl} & point & 63.5 & 66.6 \\

SegGCN~\cite{lei2020seggcn} & point & - & 58.9 \\ 

RandLA-Net~\cite{hu2020randla} & point & - & 64.5 \\

KPConv~\cite{thomas2019kpconv} & point & 69.2 & 68.6 \\

JSENet~\cite{hu2020jsenet} & point & - & 69.9 \\

FusionNet~\cite{zhang2020deep} & point & - & 68.8 \\

PointTransformer~\cite{zhao2020point} & point & 70.6 & - \\

SparseConvNet~\cite{3DSemanticSegmentationWithSubmanifoldSparseConvNet} & voxel & 69.3 & 72.5 \\

MinkowskiNet~\cite{choy20194d} & voxel & 72.2 & 73.6 \\

\specialrule{0em}{2pt}{0pt}
\hline
\specialrule{0em}{2pt}{0pt}

Ours & point & \textbf{74.3} & \textbf{73.7} \\




\bottomrule                         

\end{tabular}
\end{center}
\vspace{-0.5cm}
\caption{Results on ScanNetv2 for semantic segmentation. More results and analysis are included in the supplementary file.}
\label{table:exp_scannet}
\vspace{-0.5cm}
\end{table}

\subsection{Results}
\label{exp:results}
We make comparisons with recent state-of-the-art semantic segmentation methods. Tables~\ref{table:exp_s3dis} and \ref{table:exp_scannet} show the results on S3DIS and ScanNetv2 datasets. Our method achieves state-of-the-art performance on both challenging datasets. On S3DIS, ours outperforms others significantly, even higher than Point Transformer~\cite{zhao2020point} by $1.6\%$ mIoU. On ScanNetv2, the validation mIoU of our method surpasses others including voxel-based methods, with a gap of $2.1\%$ mIoU. On the test set, ours achieves slightly higher results than MinkowskiNet~\cite{choy20194d}. The potential reason may be the points in ScanNetv2 are relatively sparse. So the loss of accurate position in voxelization is negligible for voxel-based methods. But on S3DIS where points are denser, our method outperforms MinkowskiNet with a huge gap, \ie, $6.6\%$ mIoU. Also, ours outperforms MinkowskiNet by $2.1\%$ mIoU on the validation set and is much more robust than MinkowskiNet when encountering various perturbations in testing, as shown in Table~\ref{table:robust}. Notably, it is the first time for the point-based methods to achieve higher performance compared with voxel-based methods on ScanNetv2.

Also, in Table~\ref{tab:exp_shapenet}, to show the generalization ability, we also make comparison on ShapeNetPart~\cite{chang2015shapenet} for the task of part segmentation. Our method outperforms previous ones and achieves new state of the art in terms of both category mIoU and instance mIoU. Although the instance mIoU of ours is comparable to Point Transformer, ours outperforms Point Transformer by a large margin in category mIoU.

\begin{table}
    \centering
    \tabcolsep=0.15cm
    {
        \begin{footnotesize}
        \begin{tabular}{ l | c |  c }
            \toprule
            Method & Cat. mIoU & Ins. mIoU \\

            \specialrule{0em}{0pt}{1pt}
            \hline
            \specialrule{0em}{0pt}{1pt}
            
            
            
            PointNet~\cite{qi2017pointnet} & 80.4 & 83.7 \\
            PointNet++~\cite{qi2017pointnet++} & 81.9 & 85.1 \\
            PCNN~\cite{atzmon2018point} & 81.8 & 85.1 \\
            SpiderCNN~\cite{xu2018spidercnn} & 82.4 & 85.3 \\
            SPLATNet~\cite{su2018splatnet} & 83.7 & 85.4 \\ 
            DGCNN~\cite{wang2019dynamic} & 82.3 & 85.2 \\
            SubSparseCNN~\cite{3DSemanticSegmentationWithSubmanifoldSparseConvNet} & 83.3 & 86.0 \\
            PointCNN~\cite{li2018pointcnn} & 84.6 & 86.1 \\
            PointConv~\cite{wu2019pointconv} & 82.8 & 85.7 \\
            Point2Sequence~\cite{liu2019point2sequence} & - & 85.2 \\ 
            PVCNN~\cite{liu2019pvcnn} & - & 86.2 \\
            RS-CNN~\cite{liu2019relation} & 84.0 & 86.2 \\
            KPConv~\cite{thomas2019kpconv} & 85.0 & 86.2 \\
            InterpCNN~\cite{mao2019interpolated} & 84.0 & 86.3 \\ 
            DensePoint~\cite{liu2019densepoint} & 84.2 & 86.4 \\ 
            PAConv~\cite{xu2021paconv} & 84.6 & 86.1 \\
            PointTransformer~\cite{zhao2020point} & 83.7 & \textbf{86.6} \\
            
            \specialrule{0em}{0pt}{1pt}
            \hline
            \specialrule{0em}{0pt}{1pt}
            
            Ours & \textbf{85.1} & \textbf{86.6} \\
            
            \bottomrule                                   
        \end{tabular}
        \end{footnotesize}
    }    
    \vspace{-0.1cm}
    \caption{Results on ShapeNetPart for part segmentation.}
    \label{tab:exp_shapenet}   
\vspace{-0.2cm}
\end{table}

\subsection{Ablation Study}
\label{exp:ablation}

\begin{table}[!t]
    \centering
    \tabcolsep=0.15cm
    {
        \begin{footnotesize}
        \begin{tabular}{ c |  c  c  c  c  | c c }
            \toprule
            ID & PointEmb & Aug & cRPE & Stratified & S3DIS & ScanNet \\

            \specialrule{0em}{0pt}{1pt}
            \hline
            \specialrule{0em}{0pt}{1pt}
            
            
            
            \uppercase\expandafter{\romannumeral1} & & & & & 56.8 & 56.8 \\ 
            
            \uppercase\expandafter{\romannumeral2} & \Checkmark & & & & 61.3 & 69.6\\ 
            
            \uppercase\expandafter{\romannumeral3} & \Checkmark &  \Checkmark & & & 67.2 & 70.6 \\ 
            
            \uppercase\expandafter{\romannumeral4} & \Checkmark &  \Checkmark & \Checkmark & & 70.1 & 72.5 \\
            
            \uppercase\expandafter{\romannumeral5} & \Checkmark &  \Checkmark & \Checkmark & \Checkmark & \textbf{72.0} & \textbf{73.7} \\
            
            \specialrule{0em}{0pt}{1pt}
            \hline
            \specialrule{0em}{0pt}{1pt}
            
            
            \uppercase\expandafter{\romannumeral6} &  &  \Checkmark & \Checkmark & \Checkmark &  70.0 & 69.7 \\
            
            \uppercase\expandafter{\romannumeral7} & \Checkmark &   & \Checkmark & \Checkmark &  66.1 & 72.3 \\
            
            \uppercase\expandafter{\romannumeral8} & \Checkmark &  \Checkmark &  & \Checkmark & 68.0 & 71.4 \\
            
            \bottomrule                                   
        \end{tabular}
        \end{footnotesize}
    }    
    \vspace{-0.1cm}
    \caption{Ablation study. \textbf{PointEmb}: First-layer Point Embedding. \textbf{Aug}: Data Augmentation. \textbf{cRPE}: contextual Relative Position Encoding. \textbf{Stratified}: Stratified Transformer Block. Metric: mIoU. }
    \label{tab:ablation}   
\vspace{-0.4cm}
\end{table}

We conduct extensive ablation studies to verify the effectiveness of each component in our method, and show results in Table~\ref{tab:ablation}. To make our conclusions more convincing, we make evaluations on both S3DIS and ScanNetv2 datasets. From Exp.\uppercase\expandafter{\romannumeral1} to \uppercase\expandafter{\romannumeral5}, we add one component each time. Also, from Exp.\uppercase\expandafter{\romannumeral6} to \uppercase\expandafter{\romannumeral8}, we make double verification by removing each component from the final model, \ie, Exp.\uppercase\expandafter{\romannumeral5}.

\vspace{-0.3cm}
\paragraph{Stratified Transformer.} 
In Table~\ref{tab:ablation}, comparing Exp.\uppercase\expandafter{\romannumeral4} and \uppercase\expandafter{\romannumeral5}, we notice that with the stratified strategy, the model improves with $1.9\%$ mIoU on S3DIS and $1.2\%$ mIoU on ScanNetv2. Combining the visualizations in Fig.~\ref{fig:erf}, we note that the stratified strategy is able to enlarge the effective receptive field and boost the performance. Besides, we also show the effect when setting different downsample scales, \ie, $4$, $8$ and $16$, in the supplementary file.


\vspace{-0.3cm}
\paragraph{First-layer Point Embedding.}
We compare Exp.\uppercase\expandafter{\romannumeral1} with \uppercase\expandafter{\romannumeral2}, and find the model improves by a large margin with first-layer point embedding. Also, we compare Exp.\uppercase\expandafter{\romannumeral6} and \uppercase\expandafter{\romannumeral5}, where the model gets $2.0\%$ mIoU gain on S3DIS and $4.0\%$ mIoU gain on ScanNetv2 with the equipment of first-layer point embedding. This minor modification in the architecture brings considerable benefit.

To further explore the role of local aggregation in first-layer point embedding, we compare different ways of local aggregation with linear projection in Table~\ref{table:exp_local_agg}. Obviously, all listed local aggregation methods are better than linear projection for the first-layer point embedding.

\begin{table}[!t]
    \centering
    \tabcolsep=0.15cm
    {
        \begin{footnotesize}
        \begin{tabular}{  c | c | c | c | c | c }
            \toprule
            Method & Linear & PointTrans block & Max pool & Avg pool & KPConv \\
            
            \specialrule{0em}{0pt}{1pt}
            \hline
            \specialrule{0em}{0pt}{1pt}
            
            mIoU & 68.9 & 69.7 & 70.3 & 71.0 & 72.0 \\

            $\Delta$ & - & +0.8 & +1.4 & +2.1 & +3.1 \\

            
            \bottomrule                                   
        \end{tabular}
        \end{footnotesize}
    }    
    \vspace{-0.3cm}
    \caption{Comparison among different ways of first-layer point embedding on S3DIS. PointTrans block: Point Transformer block.}
    \label{table:exp_local_agg}
\vspace{-0.1cm}
\end{table}

\begin{table}[!t]
    \centering
    \tabcolsep=0.15cm
    {
        \begin{footnotesize}
        \begin{tabular}{  c | c | c | c | c | c | c | c | c | c}
            \toprule
            Query & &  & \Checkmark &  &  & \Checkmark & \Checkmark & & \Checkmark  \\

            Key & & &  & \Checkmark &  & \Checkmark & & \Checkmark & \Checkmark \\

            Value & &  &  &  & \Checkmark & & \Checkmark & \Checkmark & \Checkmark  \\

            MLP & \Checkmark &  &  &  & & & & &  \\

            \specialrule{0em}{0pt}{1pt}
            \hline
            \specialrule{0em}{0pt}{1pt}
            
            mIoU & 68.0 & 68.0 & 70.2 & 70.5 & 70.8 & 70.8 & 71.0 & 70.8 & 72.0 \\

            
            \bottomrule                                   
        \end{tabular}
        \end{footnotesize}
    }    
    \vspace{-0.3cm}
    \caption{Ablation study on cRPE. We evaluate on the S3DIS dataset. \textbf{Query, Key and Value}: applying the cRPE on the corresponding features to get positional bias. \textbf{MLP}: MLP-based relative position encoding.}
    \label{tab:ablation_crpe}
\vspace{-0.4cm}
\end{table}

\begin{figure*}
	\centering
    \begin{minipage}  {0.19\linewidth}
        \centering
        \includegraphics [width=1\linewidth,height=0.5\linewidth]
        {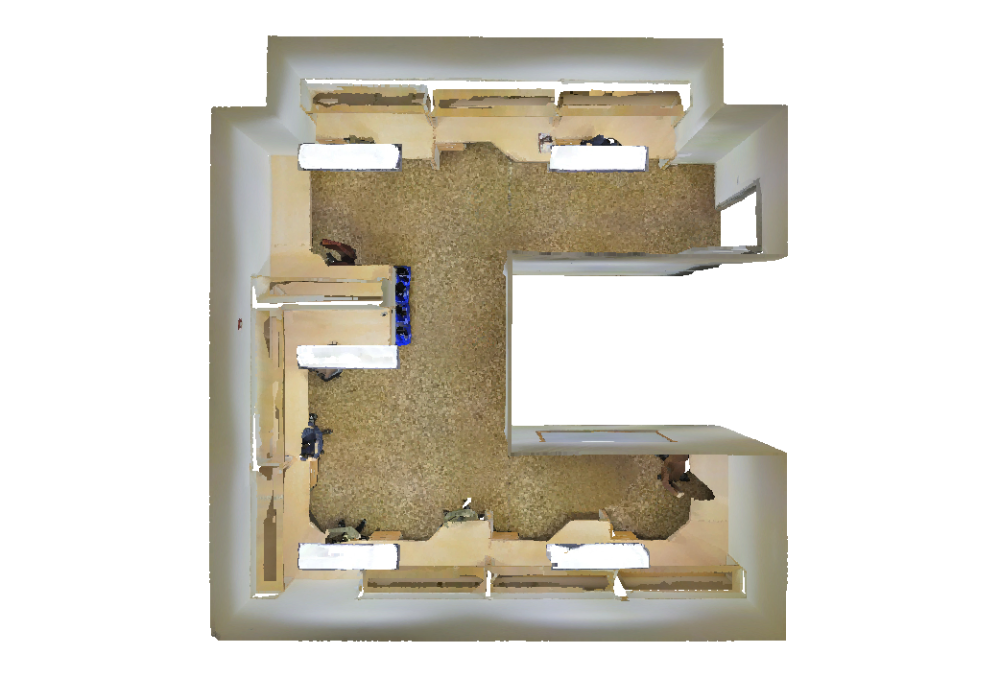}
    \end{minipage}      
     \begin{minipage}  {0.19\linewidth}
        \centering
        \includegraphics [width=1\linewidth,height=0.5\linewidth]
        {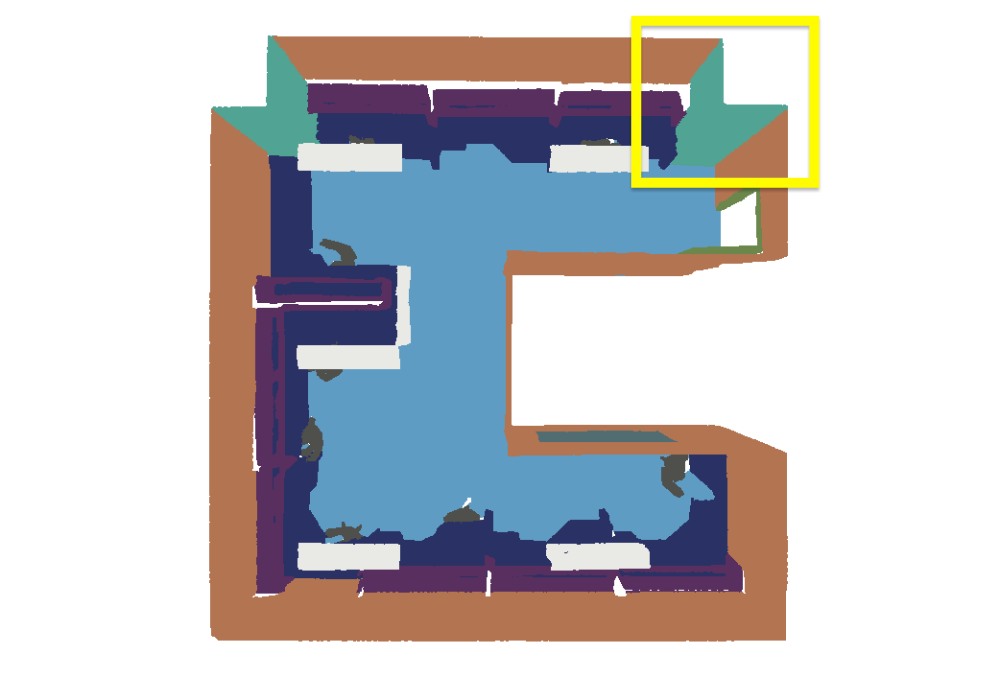}
    \end{minipage} 
    \begin{minipage}  {0.19\linewidth}
        \centering
        \includegraphics [width=1\linewidth,height=0.5\linewidth]
        {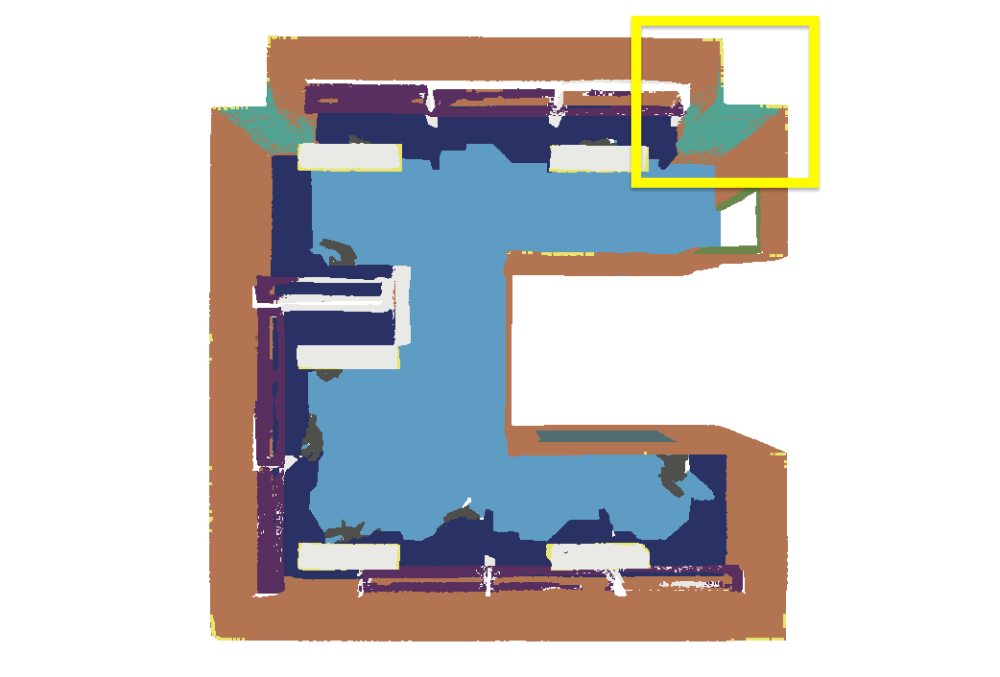}
    \end{minipage}      
     \begin{minipage}  {0.19\linewidth}
        \centering
        \includegraphics [width=1\linewidth,height=0.5\linewidth]
        {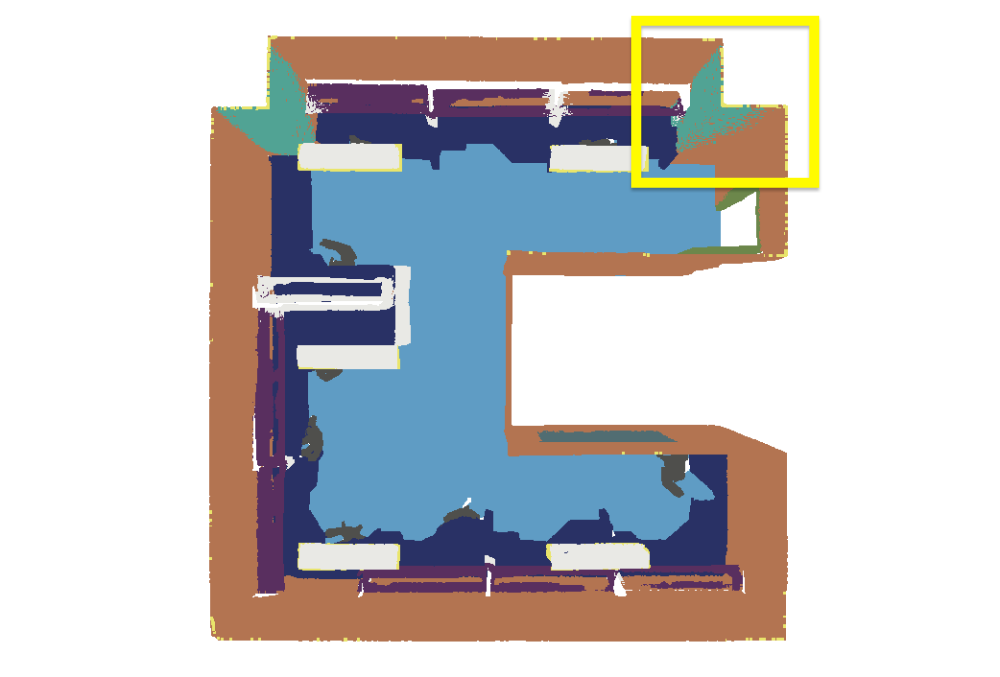}
    \end{minipage} 
     \begin{minipage}  {0.19\linewidth}
        \centering
        \includegraphics [width=1\linewidth,height=0.5\linewidth]
        {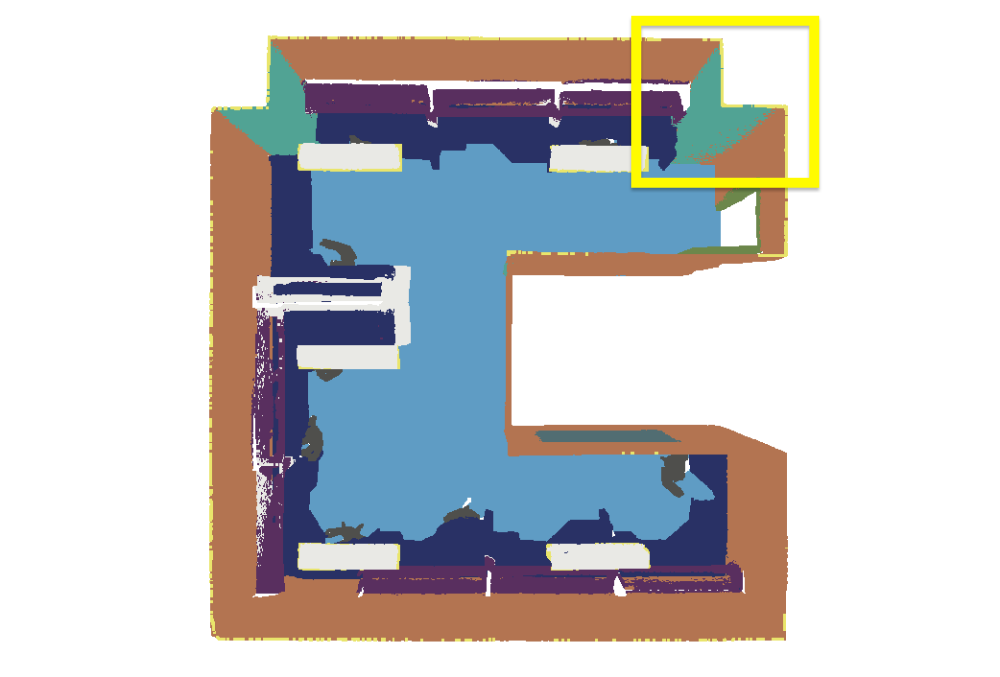}
    \end{minipage} 
	 
	 
    \begin{minipage}  {0.19\linewidth}
        \centering
        \includegraphics [width=1\linewidth,height=0.5\linewidth]
        {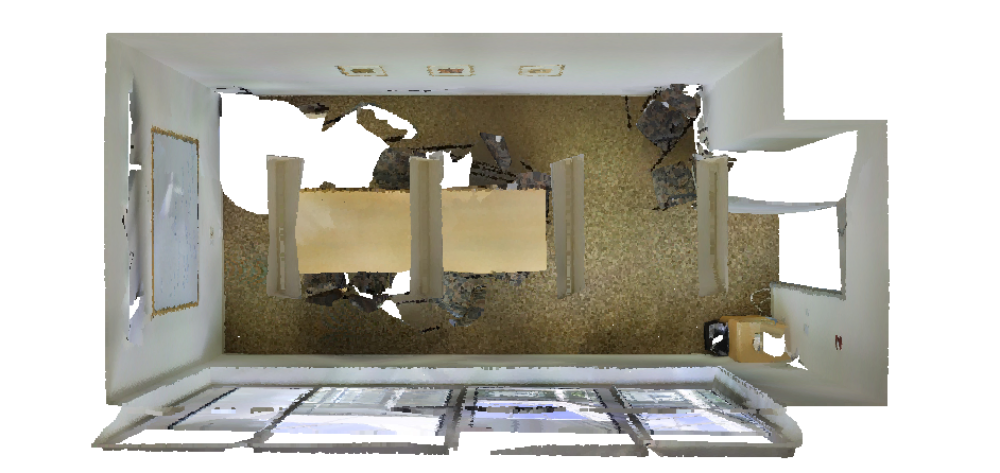}
    \end{minipage}      
     \begin{minipage}  {0.19\linewidth}
        \centering
        \includegraphics [width=1\linewidth,height=0.5\linewidth]
        {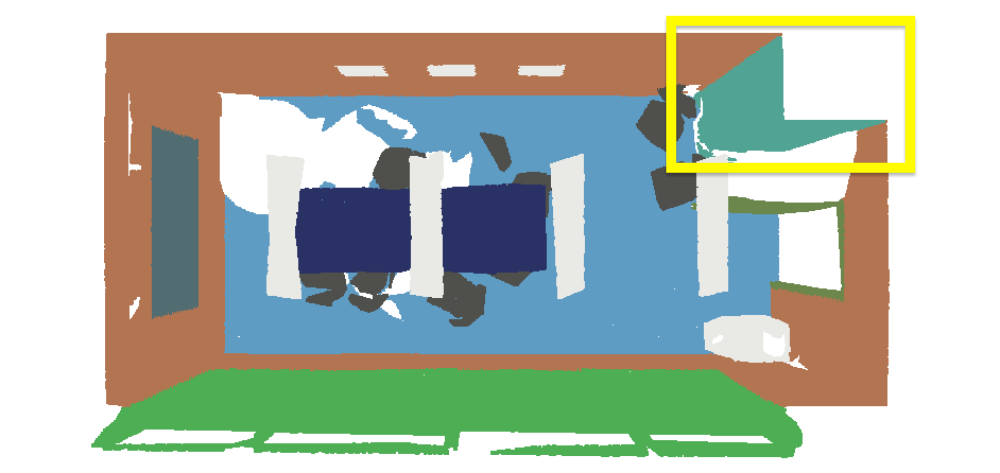}
    \end{minipage} 
    \begin{minipage}  {0.19\linewidth}
        \centering
        \includegraphics [width=1\linewidth,height=0.5\linewidth]
        {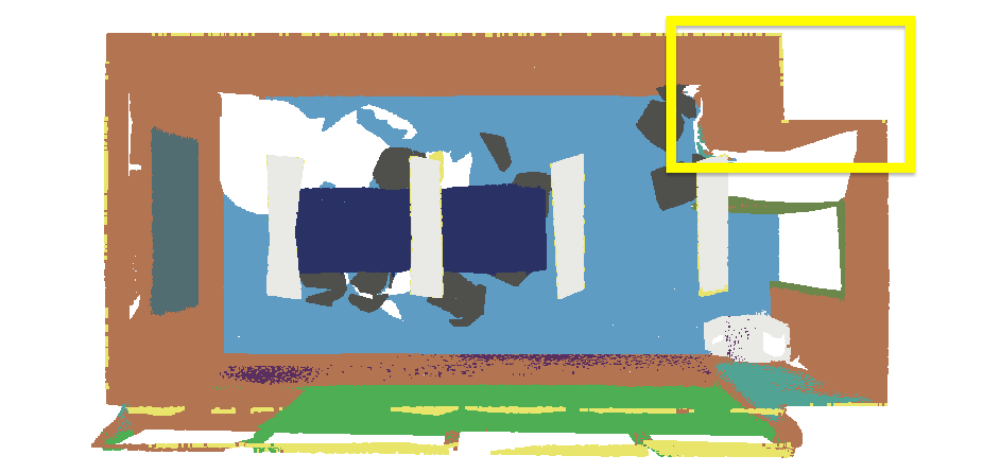}
    \end{minipage}      
     \begin{minipage}  {0.19\linewidth}
        \centering
        \includegraphics [width=1\linewidth,height=0.5\linewidth]
        {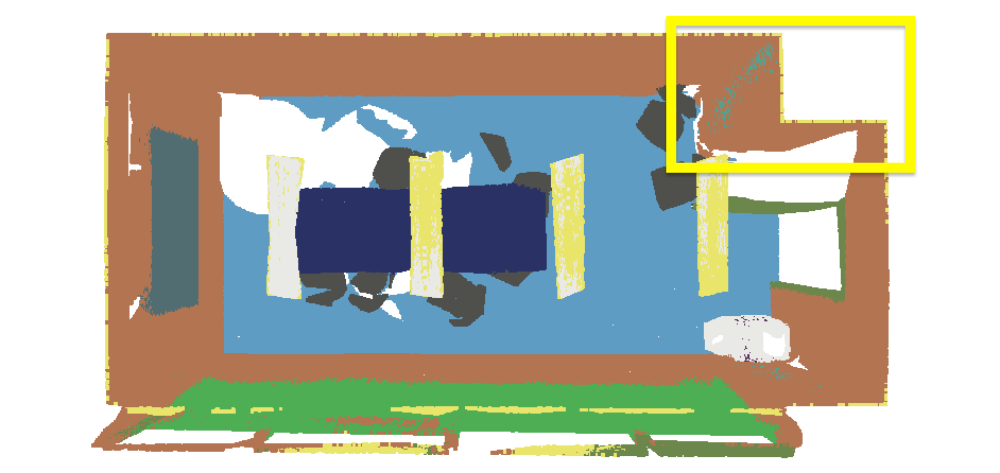}
    \end{minipage} 
     \begin{minipage}  {0.19\linewidth}
        \centering
        \includegraphics [width=1\linewidth,height=0.5\linewidth]
        {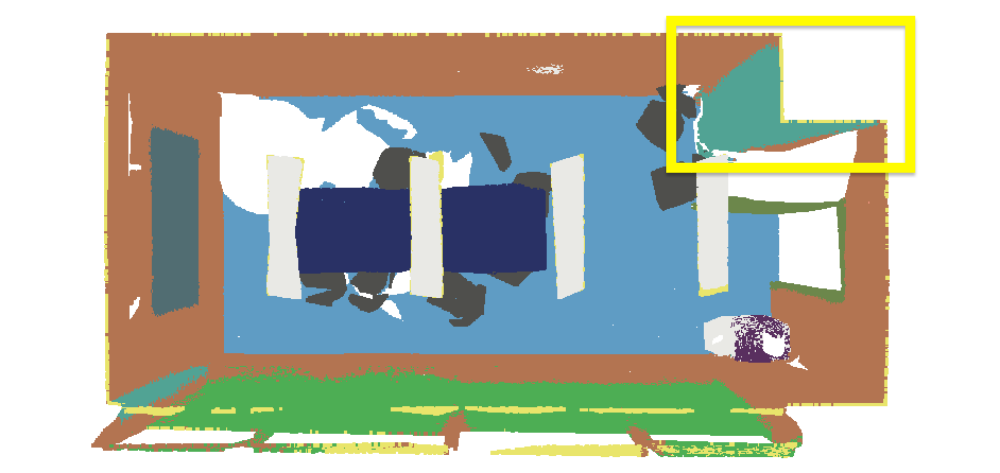}
    \end{minipage} 
    
	 
	 
    \begin{minipage}  {0.19\linewidth}
        \centering
        \includegraphics [width=1\linewidth,height=0.5\linewidth]
        {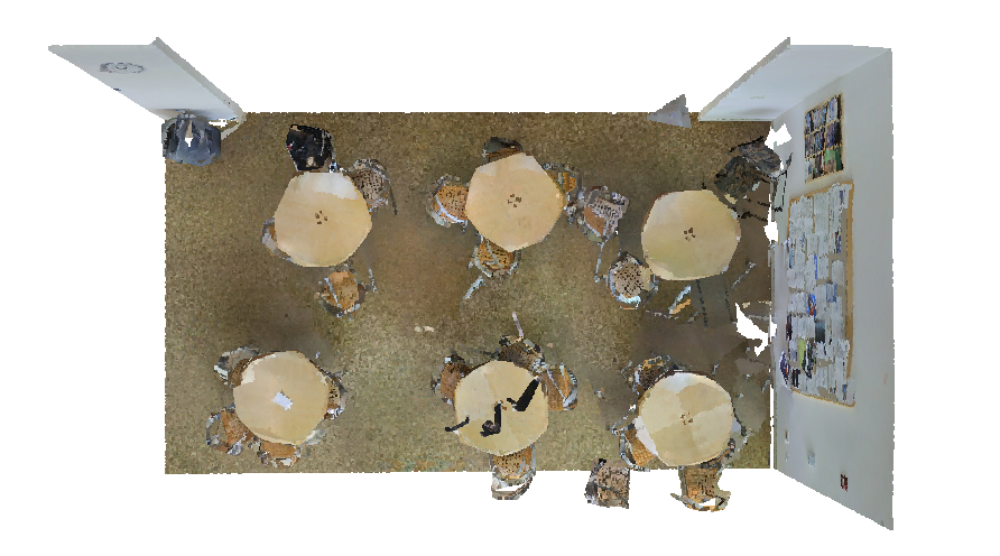}\\\footnotesize{Input}
    \end{minipage} 
     \begin{minipage}  {0.19\linewidth}
        \centering
        \includegraphics [width=1\linewidth,height=0.5\linewidth]
        {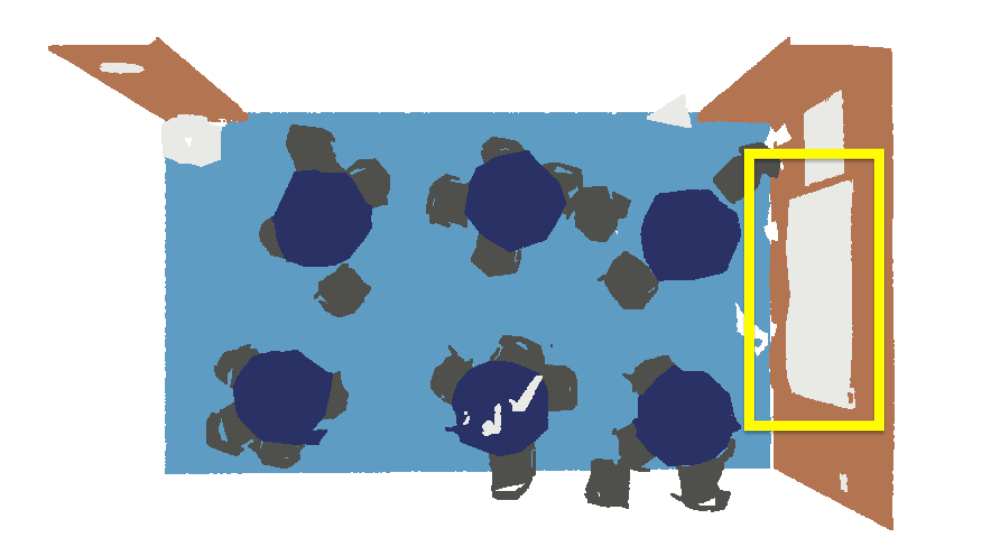}\\\footnotesize{Ground Truth}
    \end{minipage} 
    \begin{minipage}  {0.19\linewidth}
        \centering
        \includegraphics [width=1\linewidth,height=0.5\linewidth]
        {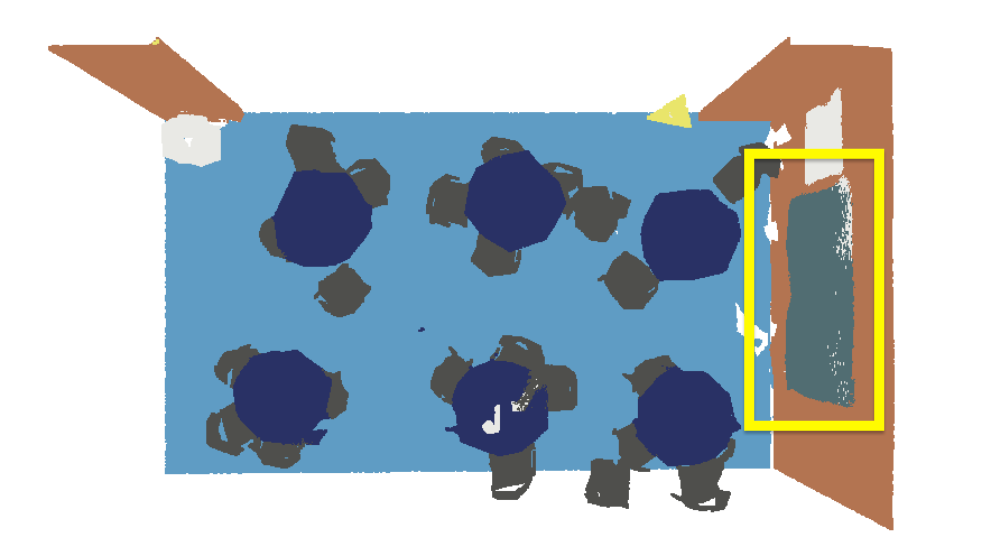}\\\footnotesize{Point Transformer}
    \end{minipage}      
     \begin{minipage}  {0.19\linewidth}
        \centering
        \includegraphics [width=1\linewidth,height=0.5\linewidth]
        {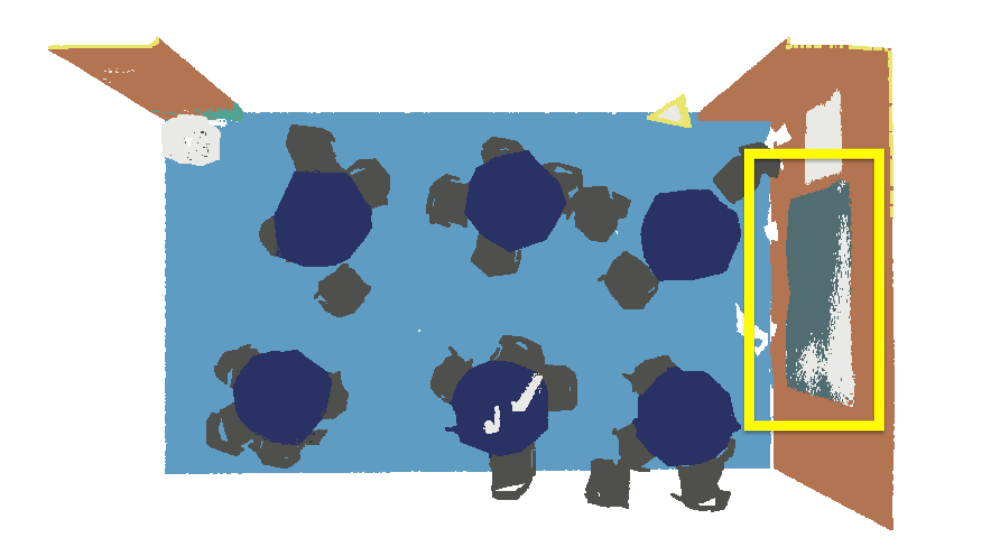}\\\footnotesize{Ours (w/o stratified)}
    \end{minipage} 
     \begin{minipage}  {0.19\linewidth}
        \centering
        \includegraphics [width=1\linewidth,height=0.5\linewidth]
        {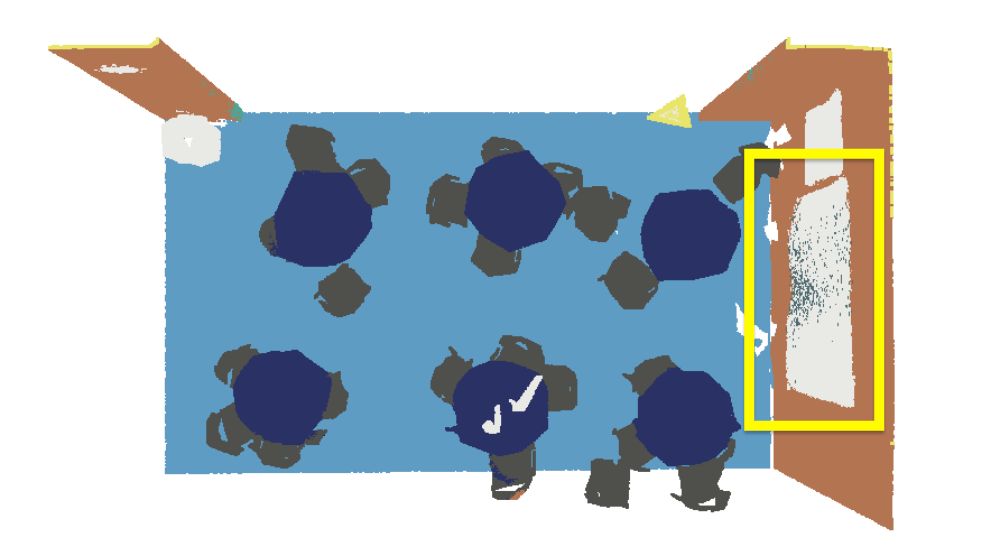}\\\footnotesize{Ours (w/ stratified)}
    \end{minipage} 
	 
	 \vspace{0.1cm}
	 
    \begin{minipage}  {0.04\linewidth}
        \centering
        \includegraphics [width=0.5\linewidth,height=0.5\linewidth]
        {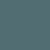}
    \end{minipage}\footnotesize board
    \begin{minipage}  {0.04\linewidth}
        \centering
        \includegraphics [width=0.5\linewidth,height=0.5\linewidth]
        {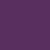}
    \end{minipage}\footnotesize bookcase
    \begin{minipage}  {0.04\linewidth}
        \centering
        \includegraphics [width=0.5\linewidth,height=0.5\linewidth]
        {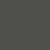}
    \end{minipage}\footnotesize chair
    \begin{minipage}  {0.04\linewidth}
        \centering
        \includegraphics [width=0.5\linewidth,height=0.5\linewidth]
        {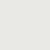}
    \end{minipage}\footnotesize clutter
    \begin{minipage}  {0.04\linewidth}
        \centering
        \includegraphics [width=0.5\linewidth,height=0.5\linewidth]
        {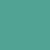}
    \end{minipage}\footnotesize column
    \begin{minipage}  {0.04\linewidth}
        \centering
        \includegraphics [width=0.5\linewidth,height=0.5\linewidth]
        {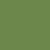}
    \end{minipage}\footnotesize door
    \begin{minipage}  {0.04\linewidth}
        \centering
        \includegraphics [width=0.5\linewidth,height=0.5\linewidth]
        {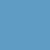}
    \end{minipage}\footnotesize floor
    \begin{minipage}  {0.04\linewidth}
        \centering
        \includegraphics [width=0.5\linewidth,height=0.5\linewidth]
        {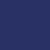}
    \end{minipage}\footnotesize table
    \begin{minipage}  {0.04\linewidth}
        \centering
        \includegraphics [width=0.5\linewidth,height=0.5\linewidth]
        {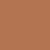}
    \end{minipage}\footnotesize wall
    \begin{minipage}  {0.04\linewidth}
        \centering
        \includegraphics [width=0.5\linewidth,height=0.5\linewidth]
        {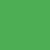}
    \end{minipage}\footnotesize window
	 
    \vspace{-0.2cm}     
    \caption{Visual comparison between Point Transformer, our baseline model (w/o stratified) and our proposed Stratified Transformer.}
    \label{fig:vis_comparison}
\vspace{-0.4cm}
\end{figure*}

\vspace{-0.3cm}
\paragraph{Contextual Relative Position Encoding.}
From Exp.\uppercase\expandafter{\romannumeral3} to \uppercase\expandafter{\romannumeral4}, the performance increases by $2.9\%$ mIoU on S3DIS and $1.9\%$ mIoU on ScanNetv2 after using cRPE. Moreover, when also using the stratified Transformer, the model still improves with $4.0\%$ mIoU gain on S3DIS and $2.3\%$ gain on ScanNetv2 equipped with cRPE, through the comparison between Exp.\uppercase\expandafter{\romannumeral8} and \uppercase\expandafter{\romannumeral5}.

Further, we testify the contribution of applying cRPE on each of the query, key or value features. Table~\ref{tab:ablation_crpe} shows that applying cRPE in either feature can make improvement. When applying cRPE on query, key and value simultaneously, the model achieves the best performance.

In addition, we compare our approach with the MLP-based method as mentioned in Sec.~\ref{sec:crpe}. As shown in Table~\ref{tab:ablation_crpe}, we find the MLP-based method (the first column) actually makes no difference with the model without any position encoding (the second column). Combining the visualization in Fig.~\ref{fig:rpe}, we conclude that the relative position information purely based on \textit{xyz} coordinates is not helpful, since input point features to the network have already incorporated the \textit{xyz} coordinates. In contrast, cRPE is based on both \textit{xyz} coordinates and contextual features.

\begin{table}
\begin{center}
\begin{footnotesize}
\begin{tabular}{l | c | c | c | c}
\hline
\toprule 
Method & w/ shift & w/o shift & shift (original) & shift (large) \\

\specialrule{0em}{2pt}{0pt}
\hline
\specialrule{0em}{2pt}{0pt}

vanilla & 70.1 & 69.4 & N/A & N/A \\
\specialrule{0em}{2pt}{0pt}
\hline
\specialrule{0em}{2pt}{0pt}

stratified & 72.0 & 70.1 & 71.0 & 70.3 \\

\bottomrule
\end{tabular}
\end{footnotesize}
\end{center}
\vspace{-0.5cm}
\caption{Ablation study on shifted window. shift (original): apply shifted window only on original windows. shift (large): apply shifted window only on large windows. vanilla: vanilla version Transformer block. }
\label{table:shifted_window}
\vspace{-0.1cm}
\end{table}

\vspace{-0.3cm}
\paragraph{Shifted Window.}
Shifted window is adopted to complement information interaction across windows. In Table~\ref{table:shifted_window}, we compare the models w/ and w/o
shifted window for both our vanilla version and Stratified Transformer on S3DIS. Evidently, shifted window is effective in our framework. Moreover, even without shifted window, Stratified Transformer still yields higher performance, \ie, $70.1\%$ mIoU, compared to the vanilla version. Also, shifting on both original and large windows is beneficial.


\begin{table}
\tabcolsep=0.17cm
\begin{center}
\begin{footnotesize}
\begin{tabular}{c | c | c | c | c | c | c}
\hline
\toprule aug type & no aug & jitter & rotate & drop color & scale & all \\
\specialrule{0em}{2pt}{0pt}
\hline
\specialrule{0em}{2pt}{0pt}

mIoU & 66.1 & 66.3 & 66.4  & 67.0 & 67.3 & 72.0\\

$\Delta$ & - & +0.2 & +0.3 & +0.9 & +1.2  & +5.9 \\

\bottomrule
\end{tabular}
\end{footnotesize}
\end{center}
\vspace{-0.5cm}
\caption{Ablation study on data augmentation evaluated on S3DIS.}
\label{table:aug}
\vspace{-0.4cm}
\end{table}

\vspace{-0.3cm}
\paragraph{Data Augmentation.}
Data augmentation plays an important role in training Transformer-based network. It is also the case in our framework as shown in Exp.\uppercase\expandafter{\romannumeral5} and \uppercase\expandafter{\romannumeral7} as well as Exp.\uppercase\expandafter{\romannumeral2} and \uppercase\expandafter{\romannumeral3}. We also investigate the contribution of each augmentation in Table~\ref{table:aug}. 





\subsection{Robustness Study}
\label{exp:robustness}


To show the anti-interference ability of our model, we measure the robustness by applying a variety of perturbations in testing. Following~\cite{xu2021paconv}, we make evaluations in aspects of permutation, rotation, shift, scale and jitter. As shown in Table~\ref{table:robust}, our method is extremely robust to various perturbations, while previous methods fluctuate drastically under these scenarios. It is notable that ours performs even better ($+0.63\%$ mIoU) with $90^{\circ}$ z-axis rotation.

Although Point Transformer also employs the self-attention mechanism, it yields limited robustness. A potential reason may be Point Transformer uses special operator designs such as ``vector self-attention" and ``subtraction relation", rather than standard multi-head self-attention. 


\begin{table}
\begin{center}
\setlength{\tabcolsep}{0.65mm}
\begin{scriptsize}
\begin{tabular}{c | c  c  c  c  c  c  c  c  c  c}
\hline
\toprule Method & None & Perm. & $90^\circ$ & $180^\circ$ & $270^\circ$ & $+0.2$ & $-0.2$ & $\times0.8$ & $\times1.2$ & jitter \\
\specialrule{0em}{2pt}{0pt}
\hline
\specialrule{0em}{2pt}{0pt}

PointNet++ & 59.75 & 59.71 & 58.15 & 57.18 & 58.19 & 22.33 & 29.85 & 56.24 & 59.74 & 59.05\\

Minkowski & 64.68 & 64.56 & 63.45 & 63.83 & 63.36 & 64.59 & 64.96 & 59.60 & 61.93 & 58.96 \\

PAConv & 65.63 & 65.64 & 61.66 & 63.48 & 61.80 & 55.81 & 57.42 & 64.20 & 63.94 & 65.12 \\

PointTrans & 70.36 & 70.45 & 65.94 & 67.78 & 65.72 & 70.44 & 70.43 & 65.73 & 66.15 & 59.67 \\

\hline
\specialrule{0em}{1pt}{1pt}
Ours & \textbf{71.96} & \textbf{72.02} & \textbf{72.59} & \textbf{72.37} & \textbf{71.86} & \textbf{71.99} & \textbf{71.93} & \textbf{70.42} & \textbf{71.21} & \textbf{72.02} \\

\bottomrule
\end{tabular}
\end{scriptsize}
\end{center}
\vspace{-0.5cm}
\caption{Robustness study on S3DIS. We apply the perturbations of permutation (Perm.), z-axis rotation ($90^{\circ}, 180^{\circ}, 270^{\circ}$), shifting ($\pm0.2$), scaling ($\times0.8, \times1.2$) and jitter in testing. PointTrans: Point Transformer~\cite{zhao2020point}.
}
\label{table:robust}
\vspace{-0.4cm}
\end{table}

\subsection{Visual Comparison}
\label{exp:vis_comparison}

In Fig.~\ref{fig:vis_comparison}, we visually compare Point Transformer, the baseline model and ours. It clearly shows the superiority of our method. Due to the awareness of long-range contexts, our method is able to recognize the objects highlighted with yellow box, while others fail.

\section{Conclusion}
We propose Stratified Transformer and achieve state-of-the-art results. The stratified strategy significantly enlarges the effective receptive field. Also, first-layer point embedding and an effective contextual relative position encoding are put forward. Our work answers two questions. First, it is possible to build direct long-range dependencies at low computational costs and yield higher performance. Second, standard Transformer can be applied to 3D point cloud with strong generalization ability and powerful performance.

\vspace{-0.2cm}
\section*{Acknowledgements}
\vspace{-0.1cm}
The work is supported in part by Hong Kong Research Grant Council - Early Career Scheme (Grant No. 27209621), HKU Startup Fund, HKU Seed Fund for Basic Research, and SmartMore donation fund.

{\small
\bibliographystyle{ieee_fullname}
\bibliography{egbib}
}

\end{document}